\documentclass{article}

\usepackage{iclr2027_conference,times}

\usepackage{amsmath,amsfonts,bm}

\def\eqref#1{equation~\ref{#1}}

\def\1{\bm{1}}

\DeclareMathAlphabet{\mathsfit}{\encodingdefault}{\sfdefault}{m}{sl}
\SetMathAlphabet{\mathsfit}{bold}{\encodingdefault}{\sfdefault}{bx}{n}

\usepackage{amsmath,amssymb,mathtools}
\usepackage{booktabs}
\usepackage{graphicx}
\usepackage{pdfpages}
\usepackage{microtype}
\usepackage{multirow}
\usepackage{subcaption}
\usepackage{wrapfig}
\usepackage{needspace}
\usepackage{placeins}
\usepackage{float}
\usepackage{latexml}
\usepackage[hidelinks]{hyperref}
\usepackage[capitalise,noabbrev]{cleveref}
\newcommand{\cpar}[1]{\par\vspace{2pt}\noindent\textbf{#1}\hspace{0.6em}}
\usepackage{xspace}
\usepackage{xcolor}

\newif\ificlrpreprint
\iclrpreprinttrue

\ificlrpreprint
  \iflatexml\else\iclrfinalcopy\fi
\fi

\title{Expanding the Quality--Coverage Frontier with Competitive Flow Matching}

\author{%
\parbox[t]{\dimexpr\textwidth-2\tabcolsep\relax}{\centering\normalfont
{\bfseries
\mbox{Wenxuan Peng$^{1,2}$}\hspace{0.5em}
\mbox{Richard Zhang$^{2}$}\hspace{0.5em}
\mbox{Haitian Zheng$^{2}$}\hspace{0.5em}
\mbox{Jaskirat Singh$^{2}$}\hspace{0.5em}
\mbox{Tsai-Shien Chen$^{2}$}\\[0.15em]
\mbox{Eli Shechtman$^{2}$}\hspace{0.5em}
\mbox{Bharath Hariharan$^{1}$}\hspace{0.5em}
\mbox{Hadar Averbuch-Elor$^{1}$}\hspace{0.5em}
\mbox{Zongze Wu$^{2}$}}\\[0.4em]
\mbox{$^{1}$Cornell University}\quad
\mbox{$^{2}$Adobe Research}
\\[0.4em]
{\small Project page: \url{https://lilydaytoy.github.io/CFM/}}
}
}

\ificlrpreprint
  \AtBeginDocument{%
    \setlength{\abovedisplayskip}{10pt plus 2pt}%
    \setlength{\belowdisplayskip}{10pt plus 2pt}%
    \setlength{\abovedisplayshortskip}{6pt plus 2pt}%
    \setlength{\belowdisplayshortskip}{8pt plus 2pt}%
  }
\fi

\begin{document}

\ificlrpreprint
  \raggedbottom
\fi

\maketitle
\ificlrpreprint
  \lhead{\textit{Preprint}}
\fi

\iflatexml
\begin{figure}[t]
  \centering
  \includegraphics[width=\linewidth]{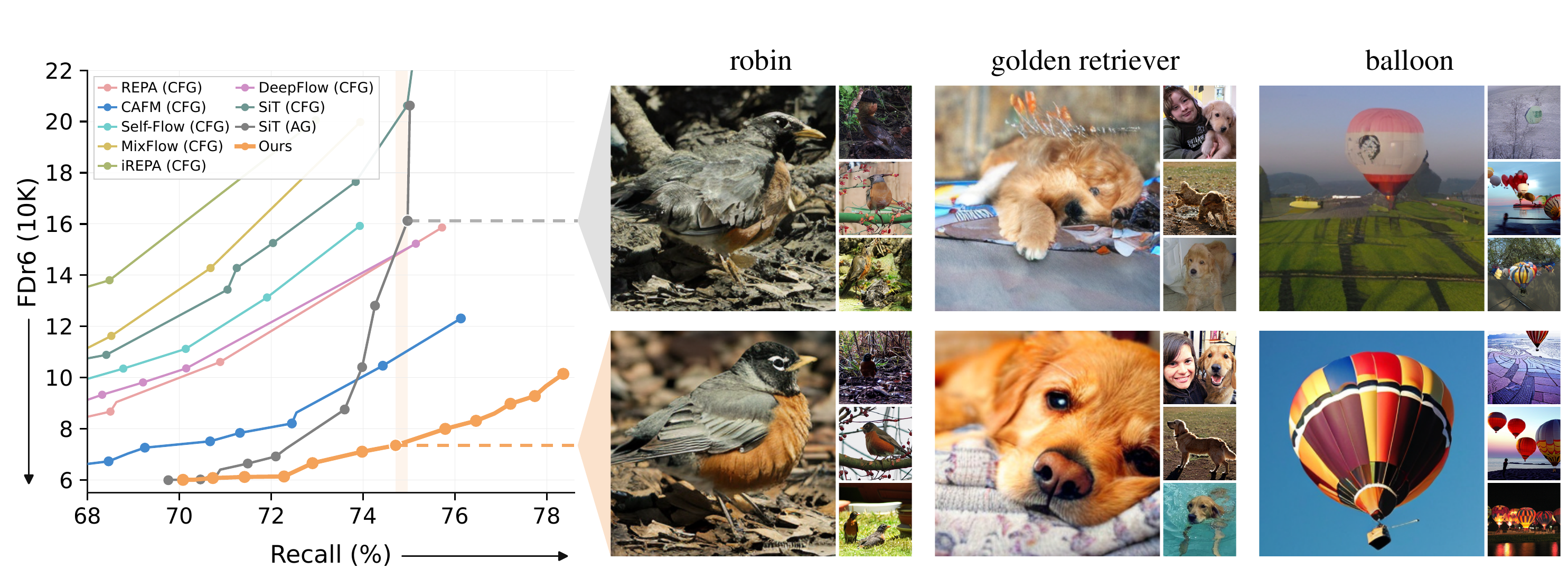}
\else
\begin{center}
\begin{minipage}{\textwidth}
  \centering
  \captionsetup{type=figure}
  \makebox[\linewidth][c]{\includegraphics[width=1.15\linewidth]{figures/teaser_guided_comparison.pdf}}
\fi
  \caption{\textbf{Our method expands the quality--coverage frontier toward high-coverage, high-quality generation.}
$\mathrm{FD}_{r,6}$--Recall Pareto frontiers reported on ImageNet-256 compare our method with recent strong flow-matching methods. Curves are obtained by sweeping each method's inference-time guidance. CFG and AG denote classifier-free guidance and AutoGuidance, respectively. Our method reaches substantially stronger performance in the high-coverage regime. This is further illustrated on the right using random samples, comparing samples obtained using the same class condition and initial noise before (top) and after (bottom) applying our approach over the SiT baseline.
}
  \label{fig:teaser}
\iflatexml
\end{figure}
\else
\end{minipage}
\end{center}
\fi

\begin{abstract}

Flow matching (FM) generators often face a quality-coverage tradeoff. Inference-time techniques such as classifier-free guidance raise quality, but at the cost of coverage. We revisit a basic property of FM -- an intermediate noisy image corresponds to multiple valid targets, which is collapsed into a single mean velocity by design. This can cover multiple modes of the distribution under ideal conditions, but is it optimal in practice? We instead investigate if directly modeling the modes of intermediate states better covers the modes of the full distribution. To do so, we augment the predictor with a randomly drawn low-dimensional code $z$, sampling $K$ candidate velocities for each training sample. We then apply the loss only to the best-matching candidate. This winner-take-all objective enables different codes to specialize and collectively cover different modes. %
Furthermore, this term naturally enables a new form of guidance, auxiliary conditioning scaling, which complements traditional guidance methods. Together, these two guidance mechanisms show that our method substantially improves the quality--coverage frontier. We validate on ImageNet-256, using two different testbeds, SiT and RAEv2. Given less than 3\% of the original SiT pretraining compute for post-training,
at the 67\% Recall point, our method reduces $\mathrm{FD}_{r,6}$ from 19.22 to 5.91. On RAEv2, at the 70\% Recall point, it reduces $\mathrm{FD}_{r,6}$ from 16.83 to 3.18.

\end{abstract}

\begin{figure}[t]
    \captionsetup{type=figure}
    \centering
    \includegraphics[width=\linewidth]{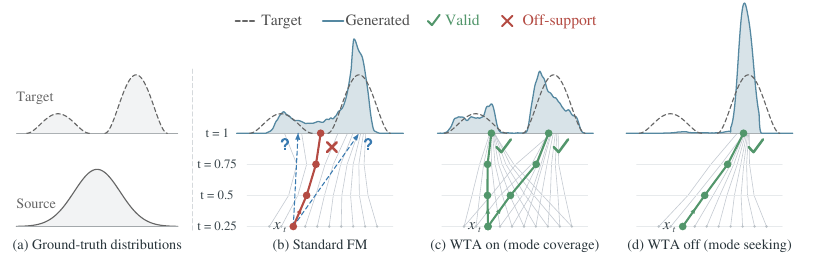}
    \caption{\textbf{A 1D toy illustrating conflicting velocity targets in flow matching.}
(a) Transport from a unimodal source to an imbalanced bimodal target.
(b) Standard FM produces off-support mass and incomplete mode coverage.
(c) Winner-take-all (WTA)-trained FM sampled with auxiliary conditioning recovers both target modes while largely removing off-support mass, even under a limited sampling budget. Two fixed codes are shown to visualize distinct code-conditioned trajectories.
(d) Sampling the same WTA-trained model without the auxiliary conditioning instead concentrates generation on the dominant mode.\protect\footnotemark}
    \label{fig:toy}
\end{figure}

\section{Introduction}
\begingroup\interfootnotelinepenalty=10000\renewcommand{\thefootnote}{\arabic{footnote}}\footnotetext{Our toy uses a simple 1D transport to isolate how finite-step integration affects the generated distribution. Standard FM performs well under sufficiently fine integration: its off-support mass decreases from 4.09\% with 4 Euler steps to 1.95\%, 1.27\%, and 0.76\% with 8, 16, and 32 steps, respectively, and further decreases to 0.51\% with 2,000 steps. Competitive FM achieves 0.41\% off-support mass with only 4 steps while covering both modes, illustrating its advantage under a limited sampling budget.}\endgroup

A successful generative model has two essential aspects: (1) \textit{high-fidelity}, where each randomly drawn sample looks realistic, and (2) \textit{broad coverage}, where the full data distribution is well-represented. Flow Matching (FM)~\citep{lipman2023flowmatching,ma2024sit} and follow-ups in architecture, representations, and training objectives~\citep{lin2026continuous, yu2024representation, shin2025deeply, chefer2026self} have yielded groundbreaking improvements in quality. In particular, to show impressive individual samples, inference-time techniques such as classifier-free guidance~\citep{ho2022cfg} and improvements~\citep{sun2024inner, chung2025cfg++, karras2024autoguidance} deliberately move towards the mode of the learned distribution, often at the expense of the tails. In Fig.~\ref{fig:teaser}, we demonstrate that while progress has indeed been made in quality (moving down the y-axis), the \textit{coverage} of the distribution (x-axis) has been underexplored. In this work, we demonstrate improvements in both quality and coverage, pushing the frontier significantly.

To explore how we might improve the quality-coverage tradeoff, we revisit flow matching through a 1D toy example (Fig~\ref{fig:toy}). Flow matching learns a velocity field, transporting samples from a source distribution to a target distribution by regressing sample-wise velocity targets. At training time, it constructs intermediate states by interpolating between randomly drawn source and target samples, training a predictor to the target. Importantly, a given intermediate state can correspond to \textit{multiple} targets, each inducing a different velocity. Under squared-error regression, these velocities are optimally combined into their conditional mean. Amazingly, with ideal modeling and sufficiently fine integration, modeling the \textit{means} can recover the full \textit{modes} of the distribution. Under practical constraints, however, generation relies on approximate learned fields and finite sampling budgets, which can leave samples off-manifold and valid alternatives underrepresented (Fig.~\ref{fig:toy}b). While post-hoc guidance can be used to bring samples back on-manifold, this brings about mode collapse and reduces mode coverage. We instead ask: rather than relying on a single marginal field (the mean) to cover all valid alternatives (the modes), can we \textbf{accommodate multiple alternatives with multiple predictions}? That is -- can we model \textit{modes} at each intermediate timestep to cover the full \textit{modes} of the distribution?

To allow the model to make multiple predictions for the same noisy state, we introduce a low-dimensional Gaussian code as auxiliary conditioning. During training, we sample multiple auxiliary codes, yielding $K$ candidate velocity predictions for the same intermediate state. Only the best-matching candidate receives the regression loss -- a winner-take-all assignment explored in
~\citep{guzman2012multiple,rupprecht2017learning}.
In our Competitive Flow Matching setup, different targets can thus shape different predictions rather than being averaged into a single compromise. 
This enables the model to produce different velocity predictions for the same noisy state, as shown in Fig.~\ref{fig:toy}(c), natively accounting for both modes.
Interestingly, turning off the auxiliary code, which reverts the model to only being able to predict a single velocity, reveals an advantageous emergent property. As shown in Fig.~\ref{fig:toy}(d), rather than going off-manifold, the generation is able to concentrate on the dominant target mode.

We validate this idea on class-conditional ImageNet-256 generation, systematically comparing against a broad range of methods in a shared setting. We first apply Competitive Flow Matching to the pretrained SiT~\citep{ma2024sit} testbed through lightweight post-training. Without inference-time guidance, our method strengthens the pretrained generator and, by varying the auxiliary-conditioning scale, exposes a
smooth, controllable quality--coverage curve rather than a single operating point.
Importantly, when combined with inference-time guidance, this learned control changes how the generator responds to guidance: generation quality improves substantially while coverage is largely preserved, and in some cases coverage and sample variation \textit{even both improve} with stronger guidance. Together, these effects substantially expand the quality--coverage
frontier, reaching high-quality, high-coverage regions that previous methods do not access (Fig.~\ref{fig:teaser}).
We further apply our technique to RAEv2~\citep{singh2026improved}, a state-of-the-art ImageNet generator operating in a pretrained representation space. WTA further improves this already strong model, establishing a new state of the art under its evaluation setting, with especially large gains at high coverage.
Finally, we analyze how WTA reshapes the trained generator itself and uncover a complementary high-fidelity behavior induced by competition.

\section{Related Work}

\paragraph{Advances in flow matching.}
Flow matching has become a powerful framework for image generation. %
Architectural improvements have made flow-based image models increasingly scalable, while representation-based methods further strengthen the features learned during generation through external alignment, self-supervision, and intermediate supervision ~\citep{yu2024representation,ICLR2026_3929a778,wang2025haste,ICLR2026_201e691e,ICLR2026_32eb1837,chefer2026self,shin2025deeply}.
More recent work has also revisited the learning signal itself, for example by training on states closer to those encountered during generation ~\citep{li2026mixflow} and constructing more stable or informative velocity targets ~\citep{yang2026stablevelocity,stoica2026pafm}.
Other advances improve generation through adversarial or contrastive supervision~\citep{lin2026continuous,Stoica_2025_ICCV} and cross-sample collaboration~\citep{mo2026group}.
These methods have pushed generation quality higher, but it is less clear whether the
gains extend to broad coverage: in our controlled evaluation, broad-coverage settings
remain substantially weaker than the strongest lower-coverage operating points.

\paragraph{Inference-time guidance.}
Beyond improving the generator itself, inference-time guidance is widely used to
improve the fidelity of individual samples.
Classifier-free guidance (CFG)~\citep{ho2022cfg} steers generation toward the modes
of the learned distribution, improving sample quality at the cost of coverage.
Subsequent methods aim to keep these gains while losing less coverage:
Inner CFG incorporates higher-order conditioning information~\citep{sun2024inner}, CFG++ attributes the degradation at strong guidance to off-manifold extrapolation and keeps the sampling update closer to the data manifold ~\citep{chung2025cfg++}.
AutoGuidance guides a strong model with a weaker version of itself, using their prediction difference to improve sample quality while better preserving variation~\citep{karras2024autoguidance}. 
These methods make the quality--coverage trade-off more favorable, but their quality gains still come primarily from moving away from broad coverage.
Rather than modifying the sampler, our work changes how flow matching learns from its
training targets, allowing guidance to reach much higher quality at broad coverage.

\paragraph{Winner-take-all learning.}
Winner-take-all (WTA) learning handles multimodal prediction by producing several
candidates per input and updating only the one closest to the target, so that
candidates specialize to different modes.
Multiple Choice Learning introduced this principle for structured prediction,
first with an ensemble of models and later with multiple heads of a single
network~\citep{guzman2012multiple,lee2016stochastic,rupprecht2017learning}.
IMLE applies the same nearest-candidate matching to implicit generative modeling,
matching each data point to its nearest generated sample~\citep{li2018imle}, and
conditional IMLE extends it to conditional generation by drawing several latent
codes per input~\citep{li2020cimle}.
We bring this formulation to local velocity prediction in flow matching. For the same intermediate state, multiple auxiliary-conditioned predictions compete to match the sample-wise velocity target, allowing different target velocities to be fit by different predictions.
Concurrent work on Explorative Modeling (XM)~\citep{gladstone2026xm} also trains
generative models with WTA, but its candidates come from different noise
realizations of the same clean sample and therefore lie at different noisy states,
whereas ours share a single state.
Furthermore, XM targets expressivity and training efficiency, whereas we target generation quality at
broad coverage.

\begin{figure}[!t]
  \centering
  \includegraphics[width=\linewidth]{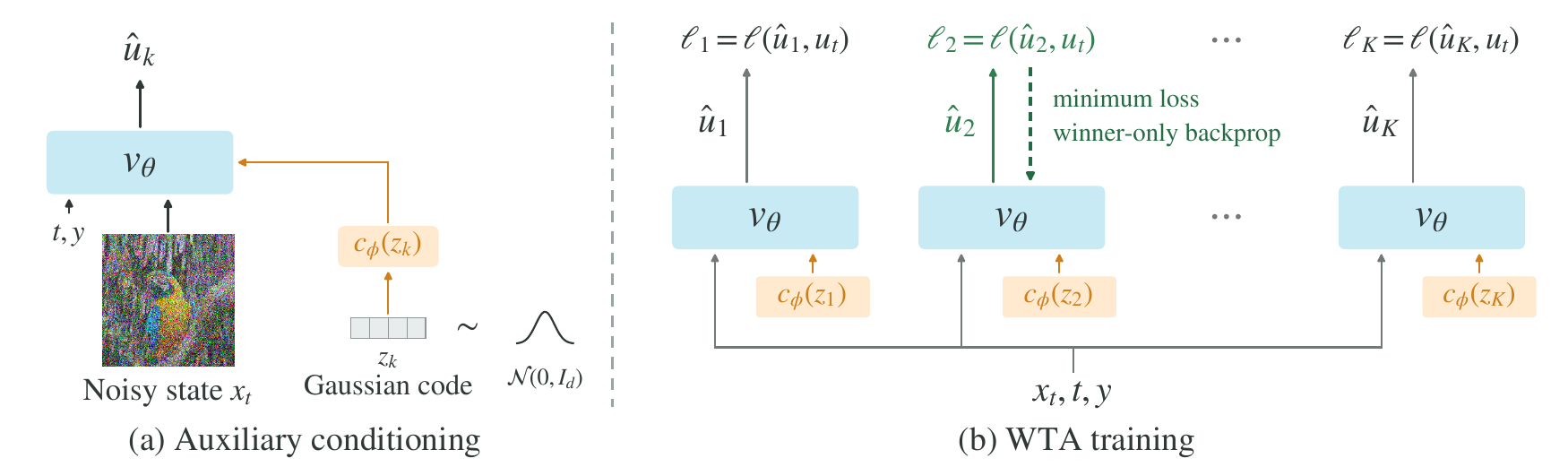}
  \captionsetup{justification=justified,singlelinecheck=false}
  \caption{\textbf{Auxiliary conditioning and WTA training.}
  (a) A Gaussian code $z_k$ is mapped through $c_\phi$ to condition the predictor $v_\theta$. (b) WTA compares $K$ predictions from one shared model against the same target.
  Only the best match backpropagates.
  }
  \label{fig:method_overview}
  \label{fig:auxiliary_architecture}
  \label{fig:aux_conditioning}
\end{figure}

\section{Method}
\paragraph{Background}
Flow matching learns a time-dependent velocity field that transports a simple noise distribution to the data distribution~\citep{lipman2023flowmatching,ma2024sit}. Given a data sample $x_0\sim p_{\mathrm{data}}$, Gaussian noise $\epsilon\sim\mathcal{N}(0,I)$, and time $t$, the training procedure constructs intermediate states $x_t$ by interpolating between data and noise points using interpolation schedule $\alpha_t, \sigma_t$, creating sample-wise velocity targets $u_t$:
\begin{equation}
x_t=\alpha_t x_0+\sigma_t\epsilon,
\qquad
u_t=\dot\alpha_t x_0+\dot\sigma_t\epsilon.
\label{eq:fm_interpolation}
\end{equation}
The velocity model $v_\theta$ is trained by squared regression, and may incorporate any extra conditioning signal $y$,
\begin{equation}
\mathcal{L}_{\mathrm{FM}}(\theta)
=
\mathbb{E}_{x_0,\epsilon,t}
\big[\lVert v_\theta(x_t,t,y)-u_t\rVert_2^2\big].
\label{eq}
\end{equation}

A given state $x_t$ and time $t$ may correspond to many different data--noise pairs, each corresponding to a different target $u_t$. Standard FM fits these
different targets with a single velocity prediction, which is their
conditional mean:
\begin{equation}
    v^*(x_t,t)
    =
    \mathbb{E}\!\left[u_t\mid x_t,t,y\right].
    \label{eq:conditional_mean}
\end{equation}

\subsection{Winner-Take-All Training}

\noindent\paragraph{Injecting low-dimensional conditioning.} To allow multiple velocity predictions for the same noisy state, we
introduce a low-dimensional Gaussian code $z$.
We use $c_\phi$ to map $z$ to a backbone-compatible auxiliary conditioning signal into $v_\theta$,
as shown in Fig.~\ref{fig:aux_conditioning}(a).
Varying $z$ enables the network to produce different velocity
predictions for the same state. Then, during training, for each noisy state $x_t$, we draw $K$ independent codes, producing $K$ candidate velocity predictions:
\begin{equation}
    \widehat u_k = v_{\theta,\phi}\big(x_t,t,y,c_\phi(z_k)\big), \text{where} \hspace{1mm} z_k \overset{\mathrm{i.i.d.}}{\sim} \mathcal{N}(0,I_d), \hspace{1mm} k\in\{1,\ldots,K\}
    \label{eq:wta_codes}
\end{equation}

As shown in Figure~\ref{fig:method_overview}(b), among the $K$ candidates, we select the one with the smallest FM regression error 
and backpropagate only through the winning prediction, which minimizes
\begin{equation}
    \mathcal{L}_{\mathrm{WTA}}(\theta,\phi)
    =
    \mathbb{E}_{x_0,y,\epsilon,t,z_{1:K}}
    \left[
    \min_{k\in\{1,\ldots,K\}}
    \left\|
    \widehat u_k-u_t
    \right\|_2^2
    \right].
    \label{eq:wta_objective}
\end{equation}

\noindent \paragraph{Implementation details.}

Though this is a general objective,  in our experiments, we leverage a pretrained model $\theta_0$ and continue training
the generator with this objective, which we refer to as
\emph{WTA post-training}.

The realization of the auxiliary conditioner $c_\phi$ depends on the
backbone architecture. We use each model's native conditioning mechanism
to implement $c_\phi$. We take SiT as a representative example.
SiT~\citep{ma2024sit} is a transformer-based flow-matching generator
operating in the latent space of a pretrained VAE. Its original
conditioning inputs, time $t$ and class label $y$, are mapped to AdaLN
modulation parameters that control the scale, shift, and gating of each
transformer block. 
To implement $c_\phi$, we map the Gaussian code $z$ through a small MLP into the same conditioning space and use a parallel AdaLN pathway to produce an auxiliary modulation of the same form. This auxiliary modulation is added to the native time--class modulation before being applied to the block. Thus, in SiT, $c_\phi$ corresponds to the code embedding together with the parallel AdaLN conditioning pathway.
We provide the corresponding implementations for other backbones, along
with exact architectural and initialization details, in the appendix.

\subsection{Inference and Guidance}
\label{sec:inference_guidance}
Once trained, there are multiple possibilities for how the auxiliary code should be used along a sampling trajectory during inference. We can  either sample one code and keep it fixed throughout the generation trajectory, or independently resample the code at each sampling step.
We empirically find that the former works better overall, particularly in the broad-coverage regime. Intuitively, this may be because keeping the same code may allow its influence on the predicted velocity field to remain consistent across sampling steps, while resampling may change this influence at every step, weakening the cumulative effect.
Therefore, unless otherwise specified, we sample $z\sim\mathcal{N}(0,I_d)$ once and keep it fixed across all sampling steps. We provide a detailed comparison with resampled codes in the appendix.

\paragraph{Guidance.}
When combining our method with inference-time guidance, there are two adjustable axes: the auxiliary-conditioning scale and the guidance strength.

\emph{Auxiliary-conditioning scale $\lambda$.}
Each auxiliary code induces a different prediction. Intuitively, varying the code is like querying different code-conditioned velocity fields. Because all codes share the same network, the WTA-trained model can be viewed as a shared base predictor, with the auxiliary conditioning on top of it producing a code-dependent response. Adjusting how strongly this conditioning influences the shared predictor provides an additional inference-time control that complements conventional guidance. We realize this adjustment by multiplying $c_\phi(z)$ by a scale $\lambda$:
\begin{equation}
v_\lambda
=
v_\theta\!\left(x_t,t,y,\lambda c_\phi(z)\right).
\label{eq:aux_scale}
\end{equation}
At $\lambda=0$, the auxiliary conditioning is removed, leaving the post-trained base model.

\emph{Guidance strength $\gamma$.}
The auxiliary-conditioning control can be combined with inference-time guidance. A common form of guidance extrapolates the generator prediction away from a guiding prediction $v_G$, which may come, for example, from an unconditional prediction (classifier-free guidance~\cite{ho2022cfg}) or a weaker model (autoguidance~\cite{karras2024autoguidance}). The guided prediction is then
\begin{equation}
v_{\lambda,\gamma}
=
v_\lambda
+
\gamma\left(v_\lambda-v_G\right).
\label{eq:combined_guidance}
\end{equation}

\section{Experiments}
We evaluate our method primarily on class-conditional ImageNet-256 generation.
We first use SiT as a controlled testbed to evaluate how our post-training affects generation quality and coverage. We then test whether these improvements generalize beyond SiT using RAEv2, and analyze the role of competition.

\subsection{SiT on ImageNet}

\paragraph{Setup.}
We use SiT as our main testbed because many recent flow-matching methods build on it,
allowing controlled comparisons within a common model family.

Our main model starts from the official pretrained SiT-XL/2 checkpoint and is
post-trained with the WTA objective for 10 epochs with $K=8$.
For guidance, we instantiate Eq.~\ref{eq:combined_guidance} with AutoGuidance
(AG)~\citep{karras2024autoguidance}: the guiding prediction $v_G$ comes from a smaller
SiT-B/2, also post-trained with our method using $K=8$ and run auxiliary-off ($\lambda=0$).
The pretrained SiT baseline uses the same AG formulation, with pretrained SiT-XL/2 and
SiT-B/2 as the generator and guiding model.

To separate differences in the learned generators from differences in sampling
heuristics, the controlled SiT comparisons, including Table~\ref{tab:guided_performance},
use a 50-step Euler ODE solver with uniformly spaced timesteps. For the methods we compare with, we re-evaluate them under this recipe when their original setup differs, retaining the guidance formulation used in their original papers and sweeping its strength.

We report $\mathrm{FD}_{r,6}$~\citep{yang2026fdloss} as our primary metric; it
aggregates Fr\'echet distances across six representation spaces, including the
Inception space underlying FID. We also report Precision (sample fidelity) and Recall
(coverage of the data distribution), with Inception Score (IS) as a supporting quality
metric and within-class DreamSim~\citep{fu2023dreamsim} distance as a supporting
diversity metric. Operating-point curves use 10K generated samples; key numerical
comparisons use 50K.

\iflatexml
\begin{figure}[t]
  \centering
  \includegraphics[width=\linewidth]{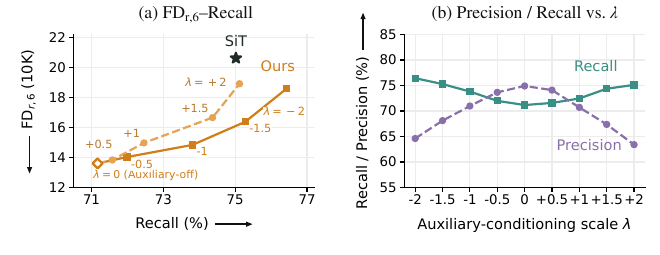}
\else
\par\medskip
\noindent\begin{minipage}{\linewidth}
  \centering
  \makebox[0.80\linewidth][l]{\hspace*{-0.03545455\linewidth}%
    \includegraphics[width=0.80\linewidth]{cfm-figure4-panel-titles-unified-20261120.pdf}}\par
  \captionsetup{type=figure}
\fi
  \caption{\textbf{Our method improves generation and enables smooth operating-point control.} (a) At $\gamma=0$, varying the auxiliary-conditioning scale $\lambda$ lowers $\mathrm{FD}_{r,6}$ relative to pretrained SiT, including at matched or higher Recall. (b) Increasing $|\lambda|$ from auxiliary-off ($\lambda=0$) trades Precision for Recall.
    }
  \label{fig:lambda_slider}
\iflatexml
\end{figure}
\else
\end{minipage}
\par\medskip
\fi

\paragraph{Varying the auxiliary conditioning scale $\lambda$.}
We first examine the effect of auxiliary conditioning by varying $\lambda$ in the post-trained generator while fixing $\gamma=0$.
Starting from the pretrained SiT checkpoint, 10 epochs of our post-training yield clear
gains, whereas continuing standard FM training does not improve generation.
As shown in Fig.~\ref{fig:lambda_slider}(a), varying $\lambda$ yields operating
points with lower $\mathrm{FD}_{r,6}$ than pretrained SiT, including at matched and
even higher Recall.
The auxiliary-conditioning scale $\lambda$ also provides a smooth inference-time knob for controlling the generator's operating point. As shown in Fig.~\ref{fig:lambda_slider}(b), the auxiliary-off setting ($\lambda=0$) lies at high
Precision, and increasing $|\lambda|$ moves the generator steadily toward higher Recall. Interestingly, we empirically observe similar overall trends for both signs of $\lambda$, with negative values generally yielding lower $\mathrm{FD}_{r,6}$.
We therefore present the negative-$\lambda$ branch in the guided frontier below and compare both signs in Appendix~\ref{sec:supp-sit-lambda-sign}.

Under the standard 50K-sample evaluation at $\gamma=0$, at matched Recall ($66.91\%$ for pretrained SiT and $66.83\%$ for our model), we reduce $\mathrm{FD}_{r,6}$ from $19.22$ to $14.94$ and
FID from $11.17$ to $7.22$. 
The auxiliary-off setting ($\lambda=0$) reaches an even lower $\mathrm{FD}_{r,6}$ of
$12.15$ (FID $4.59$).
\textit{Our method produces a stronger generator whose quality--coverage trade-off can be smoothly controlled through $\lambda$ alone.}
\begin{figure}[!htbp]
  \centering
  \iflatexml
    \includegraphics[width=\linewidth]{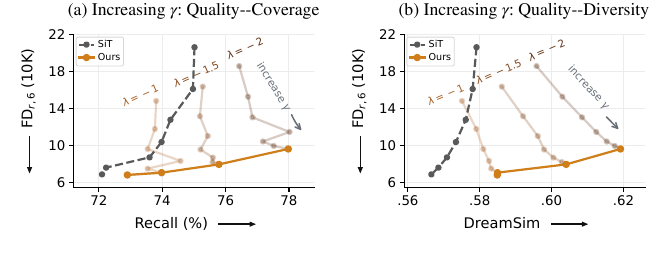}
  \else
  \makebox[0.80\linewidth][l]{\hspace*{-0.03545455\linewidth}%
    \includegraphics[width=0.80\linewidth]{figures/05_headline/guided_frontier_iclr_cvpr.pdf}}
  \fi
  \caption{\textbf{Our method enables guidance to improve generation quality at little cost to coverage.}
  Light orange curves show guidance sweeps at fixed auxiliary-conditioning scales $\lambda$, with $\gamma=0,0.25,\ldots,1.5$ from top to bottom, compared with pretrained SiT (black dashed). Thick orange curves show the resulting Pareto frontiers.
  (a) In $\mathrm{FD}_{r,6}$--Recall space, stronger guidance improves quality while largely preserving Recall for our model, unlike SiT, which loses coverage.
  (b) In $\mathrm{FD}_{r,6}$--DreamSim space, stronger guidance increases within-class diversity for our model while reducing it for SiT.}
  \label{fig:guided_frontier}
  \label{fig:dreamsim_guidance}
\end{figure}

\iflatexml\else\Needspace{11\baselineskip}\fi
\paragraph{Combining with Guidance.}
We next examine how our post-training changes the generator's response to inference-time guidance. For each fixed auxiliary-conditioning scale $\lambda$, we
sweep the guidance strength $\gamma$ to trace a guided trajectory.
Figure~\ref{fig:guided_frontier}(a) shows that pretrained SiT and our model respond very differently to guidance. For pretrained SiT, increasing guidance lowers $\mathrm{FD}_{r,6}$ but steadily reduces Recall. For our model, at a fixed $\lambda$, stronger guidance moves the trajectory mostly downward: $\mathrm{FD}_{r,6}$ decreases substantially with little or no Recall loss, while at high-coverage operating points Recall can even increase.
Combining the guided trajectories across $\lambda$ gives our Pareto frontier, which extends beyond guided SiT toward both lower
$\mathrm{FD}_{r,6}$ and higher Recall.
Figure~\ref{fig:teaser} compares this frontier with recent flow-matching
methods over their respective guidance sweeps: our method performs best overall, with its largest
advantage in the broad-coverage regime.

This advantage further extends to sample diversity. In addition to Recall, we measure within-class diversity using mean pairwise DreamSim~\citep{fu2023dreamsim} distance.
Echoing the Recall trends above, stronger guidance reduces diversity for pretrained SiT but steadily increases it for our model at high-coverage settings (Fig.~\ref{fig:dreamsim_guidance}(b)).

\begin{table}[H]
  \centering
  \setlength{\fboxsep}{1pt}
  \caption{\textbf{Guided comparison under the standard 50K-sample evaluation.}
We evaluate each method using its own reference guidance formulation, sweeping guidance strength under a shared 50-step Euler sampling protocol.
(a) Performance at each method's lowest $\mathrm{FD}_{r,6}$.
(b) Lowest $\mathrm{FD}_{r,6}$ at Recall $\approx 67\%$; light-gray Recall values mark methods whose highest measured Recall is below $67\%$.
The six raw Fr\'echet distances (including ADM FID) appear in smaller gray type;
$\mathrm{FD}_{r,6}$ averages their normalized values.}
  \label{tab:guided_performance}
  \includegraphics[width=\linewidth]{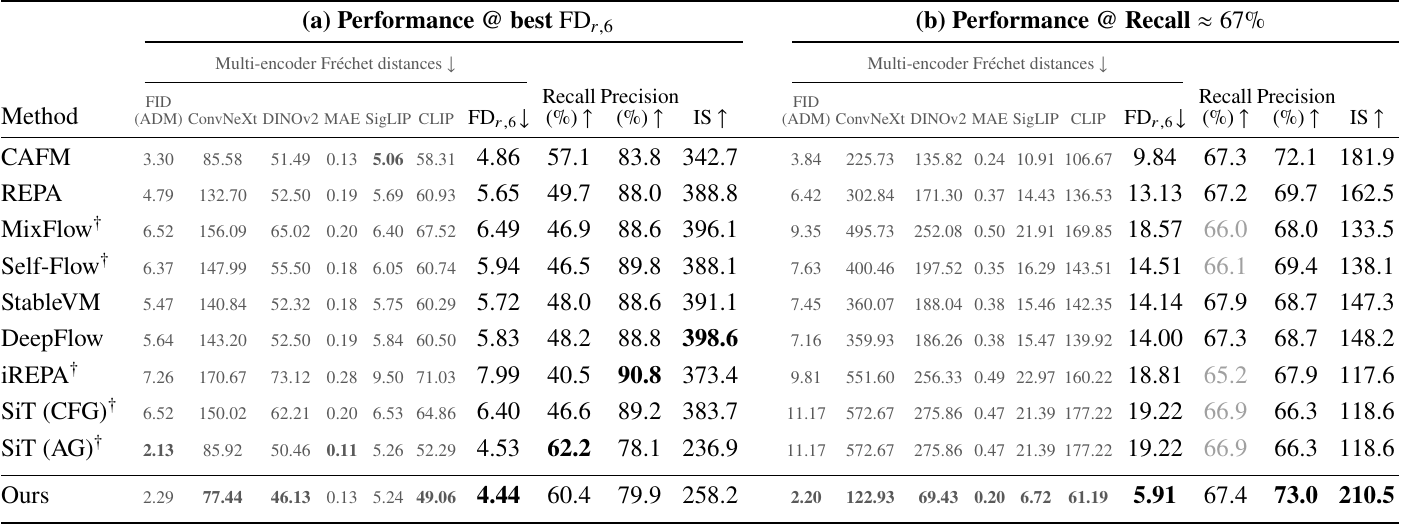}

\end{table}

Finally, Table~\ref{tab:guided_performance} reports the full set of metrics under the standard 50K-sample evaluation across methods. 
At the best overall operating point, our method achieves the lowest $\mathrm{FD}_{r,6}$ in the comparison, reaching $4.44$ (Table~\ref{tab:guided_performance}a).
The gap widens in the broad-coverage regime (Recall $\approx 67\%$,
Table~\ref{tab:guided_performance}b). Competing methods degrade sharply here, with higher
$\mathrm{FD}_{r,6}$ and lower Precision and Inception Score. Our model instead stays near its
best, reaching $\mathrm{FD}_{r,6}=5.91$ with the highest Precision and Inception Score
among the compared methods.
\textit{With our method, stronger guidance improves quality while largely preserving coverage, significantly expanding the quality--coverage frontier.}

\subsection{Extending to Stronger Generators}
We next investigate whether our post-training can further improve already strong pretrained generators. We first consider CAFM~\citep{lin2026continuous}, one of the strongest SiT-based generators in our comparison, and then extend to RAEv2~\citep{singh2026improved}, which operates in a different representation space and uses its own guidance mechanism.

\begin{figure}[H]
  \centering
  \iflatexml
    \includegraphics[width=\linewidth]{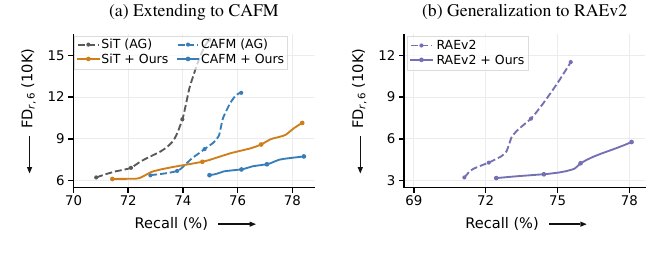}
  \else
  \makebox[0.80\linewidth][l]{\hspace*{-0.03545455\linewidth}%
    \includegraphics[width=0.80\linewidth]{figures/05_headline/stronger_generator_frontiers_iclr.pdf}}
  \fi
  \caption{\textbf{Our method further improves strong pretrained generators.}
  (a) Applying our post-training to CAFM further improves its $\mathrm{FD}_{r,6}$--Recall frontier. SiT before and after our post-training are included for reference. All models use AutoGuidance.
  (b) Applying our post-training to RAEv2 substantially improves its frontier, particularly at high Recall, using its native guidance.
  All results use 10K samples and 50-step Euler sampling.}
  \label{fig:stronger_generators}
  \label{fig:raev2_transfer}
\end{figure}

\vspace{-2.5em}
\paragraph{Further improving CAFM.}
We first consider CAFM, which substantially improves upon the plain SiT generator through adversarial training. CAFM originally reports guided results using CFG. We additionally evaluate CAFM with AG, which further improves its quality--coverage frontier over its CFG counterpart.
Importantly, our model post-trained from plain SiT still maintains a clear advantage over CAFM with AG, particularly in the high-coverage regime.
We then apply the same WTA post-training to the pretrained CAFM checkpoint. As shown in Fig.~\ref{fig:stronger_generators}(a), this yields an even stronger quality--coverage frontier than the one obtained by post-training SiT. At a high-coverage operating point (Recall $\approx 67\%$), our post-training further reduces $\mathrm{FD}_{r,6}$ from 6.08 to 4.93 on CAFM (Table~\ref{tab:stronger_generators}(a)), demonstrating that our method can build on the improvements of an already strong generator.
\vspace{-0.8em}
\paragraph{Generalization to RAEv2.}
We further test generalization across different model families and guidance mechanisms using RAEv2.
Whereas SiT operates in a VAE latent space, RAEv2 generates in the representation space
of a pretrained visual encoder and achieves state-of-the-art ImageNet generation.
RAEv2 has its own guidance mechanism, which we call RAEv2-guidance: in the notation of Sec.~\ref{sec:inference_guidance}, the guiding prediction $v_G$ comes from an intermediate head within the same forward pass. For our model, both predictions therefore share the same $\lambda$.
We post-train the pretrained RAEv2 generator for 10 epochs with $K=8$ competing predictions. Both the pretrained and post-trained models use RAEv2's native inference recipe, including its timestep shifting and guidance window. We sweep the RAEv2-guidance strength for each, and additionally vary $\lambda$ for our model.
As shown in Fig.~\ref{fig:stronger_generators}(b), our method improves RAEv2's quality--coverage frontier, with particularly large gains at high Recall. Near 70\% Recall, $\mathrm{FD}_{r,6}$ drops from 16.83 to 3.18, while the best-quality point also improves from 2.07 to 2.02 (Table~\ref{tab:stronger_generators}(b)).

\textit{Together, these results demonstrate that our post-training can further improve already strong generators across different model families and guidance mechanisms.}

\begin{table}[t]
  \centering
  \caption{\textbf{Post-training comparison under the 50K-sample evaluation.}
  We compare each generator before and after applying our post-training.
  (a) SiT and CAFM use their respective AutoGuidance configurations under a shared 50-step Euler sampling protocol.
  (b) RAEv2 follows its native inference and guidance recipe, including timestep shifting and guidance window.
  We report performance at the best $\mathrm{FD}_{r,6}$ and approximately 67\% Recall for SiT-based generators and 70\% Recall for RAEv2.}
  \label{tab:stronger_generators}
  \footnotesize
  \newcommand{\sgmetric}[1]{\makebox[0.42in][c]{#1}}
  \newcommand{\sgheader}[1]{\makebox[0.42in][c]{{\fontsize{6.4}{7.2}\selectfont #1}}}
  \setlength{\tabcolsep}{0pt}
  \renewcommand{\arraystretch}{1.04}
  \begin{tabular}{@{}l@{\hspace{6pt}}cccc@{\hspace{16pt}}cccc@{}}
    \toprule
    \multicolumn{1}{c}{\multirow{2}{*}{Model}}
      & \multicolumn{4}{c}{{\bfseries\boldmath Best $\mathrm{FD}_{r,6}$}}
      & \multicolumn{4}{c}{{\bfseries High coverage}} \\
    \cmidrule(lr){2-5}\cmidrule(lr){6-9}
      & \sgheader{$\mathrm{FD}_{r,6}\!\downarrow$}
      & \sgheader{\shortstack{Recall\\(\%) $\uparrow$}}
      & \sgheader{\shortstack{Precision\\(\%) $\uparrow$}}
      & \sgheader{IS $\uparrow$}
      & \sgheader{$\mathrm{FD}_{r,6}\!\downarrow$}
      & \sgheader{\shortstack{Recall\\(\%) $\uparrow$}}
      & \sgheader{\shortstack{Precision\\(\%) $\uparrow$}}
      & \sgheader{IS $\uparrow$} \\
    \midrule
    \multicolumn{5}{@{}l}{(a) SiT-based generators}
      & \multicolumn{4}{c}{Recall $\approx 67\%$} \\[1pt]
    SiT
      & \sgmetric{4.53} & \sgmetric{\textbf{62.2}} & \sgmetric{78.1} & \sgmetric{236.9}
      & \sgmetric{19.22} & \sgmetric{66.9} & \sgmetric{66.3} & \sgmetric{118.6} \\
    SiT + Ours
      & \sgmetric{\textbf{4.44}} & \sgmetric{60.4} & \sgmetric{\textbf{79.9}} & \sgmetric{\textbf{258.2}}
      & \sgmetric{\textbf{5.91}} & \sgmetric{67.4} & \sgmetric{\textbf{73.0}} & \sgmetric{\textbf{210.5}} \\
    \noalign{\vskip 2.25pt{\color{black!25}\hrule height 0.25pt}\vskip 2.25pt}
    CAFM
      & \sgmetric{4.85} & \sgmetric{\textbf{66.1}} & \sgmetric{74.3} & \sgmetric{226.0}
      & \sgmetric{6.08} & \sgmetric{67.4} & \sgmetric{73.9} & \sgmetric{208.8} \\
    CAFM + Ours
      & \sgmetric{\textbf{4.55}} & \sgmetric{64.6} & \sgmetric{\textbf{77.5}} & \sgmetric{\textbf{238.4}}
      & \sgmetric{\textbf{4.93}} & \sgmetric{67.5} & \sgmetric{\textbf{72.6}} & \sgmetric{\textbf{216.4}} \\
    \midrule
    \multicolumn{5}{@{}l}{(b) RAEv2}
      & \multicolumn{4}{c}{Recall $\approx 70\%$} \\[1pt]
    RAEv2
      & \sgmetric{2.07} & \sgmetric{64.72} & \sgmetric{79.39} & \sgmetric{261.7}
      & \sgmetric{16.83} & \sgmetric{69.09} & \sgmetric{52.39} & \sgmetric{111.5} \\
    RAEv2 + Ours
      & \sgmetric{\textbf{2.02}} & \sgmetric{\textbf{65.71}} & \sgmetric{\textbf{78.30}} & \sgmetric{\textbf{265.8}}
      & \sgmetric{\textbf{3.18}} & \sgmetric{70.32} & \sgmetric{\textbf{71.25}} & \sgmetric{\textbf{182.4}} \\
    \bottomrule
  \end{tabular}
\end{table}

\subsection{Analyzing the Role of Competition}
\label{sec:shared-predictor}
We next analyze how our post-training reshapes the generator and what this reveals about the role of competition. We start with the base model itself, evaluated by removing the auxiliary conditioning ($\lambda=0$), and compare it with its pretrained counterpart across generator families. We then use SiT to study how the base model changes as competition scales with $K$, and finally extend this analysis across $\lambda$.

\paragraph{Removing the auxiliary conditioning.}

\begin{table}[!htbp]
  \caption{\textbf{Base model evaluation.}
For each generator, we compare the pretrained model with its base model after
our post-training ($K=8$), evaluated at $\lambda=0$. Each model is evaluated following the inference and evaluation protocol of its original work (Appendix~\ref{sec:supp-native-evaluation}).}
  \label{tab:shared_predictor}
  \centering
   \scalebox{0.8}{\begingroup
\setlength{\tabcolsep}{2pt}
\begin{tabular*}{\linewidth}{@{\extracolsep{\fill}}l*{8}{c}@{}}
\toprule
 & \multicolumn{2}{c}{\textbf{SiT}} & \multicolumn{2}{c}{\textbf{JiT}} & \multicolumn{2}{c}{\textbf{RAE}} & \multicolumn{2}{c}{\textbf{RAEv2}}\\
\cmidrule(lr){2-3}\cmidrule(lr){4-5}\cmidrule(lr){6-7}\cmidrule(l){8-9}
Metric & Pre-trained & Ours & Pre-trained & Ours & Pre-trained & Ours & Pre-trained & Ours\\
\midrule
$\mathrm{FD}_{r,6}\downarrow$ & 15.76 & \textbf{10.59} & 15.17 & \textbf{14.15} & 4.41 & \textbf{3.64} & 3.89 & \textbf{3.52}\\
Recall (\%) $\uparrow$ & \textbf{67.65} & 62.83 & \textbf{66.46} & 62.03 & \textbf{63.80} & 58.73 & \textbf{62.08} & 55.49\\
Precision (\%) $\uparrow$ & 68.71 & \textbf{75.70} & 68.06 & \textbf{72.05} & 79.13 & \textbf{83.53} & 79.84 & \textbf{85.76}\\
Inception Score $\uparrow$ & 131.86 & \textbf{178.47} & 151.49 & \textbf{168.01} & 240.36 & \textbf{269.29} & 226.72 & \textbf{278.71}\\
\bottomrule
\end{tabular*}
\endgroup
}
\end{table}

Using $\lambda=0$ removes the auxiliary conditioning $c_\phi(z)$ and the ability to predict multiple velocities, leaving a fine-tuned version of the original base model. How does this base model compare to its pretrained version?
Table~\ref{tab:shared_predictor} evaluates this generator across representative generator families, including VAE-latent SiT~\citep{ma2024sit}, pixel-space JiT~\citep{li2025jit}, and representation-autoencoder generators RAE~\citep{zheng2026diffusion} and RAEv2~\citep{singh2026improved}. Across all four models, the base model after our post-training achieves lower $\mathrm{FD}_{r,6}$ and higher Inception Score, and shifts toward a higher-Precision, lower-Recall regime.
Interestingly, \emph{competitive post-training improves the base model itself and shapes it into a higher-fidelity generator.}

\paragraph{Scaling competition.}
\label{sec:k-scaling}

The effect on the base model is closely tied to competition. As shown in Fig.~\ref{fig:k_scaling}(a), when $K=1$, no competition occurs, and post-training leaves the base model nearly unchanged from pretrained SiT. As more predictions compete for each target, the base model shifts systematically toward higher Precision and lower $\mathrm{FD}_{r,6}$. %
This effect on $\mathrm{FD}_{r,6}$ is more pronounced with shorter sampling trajectories (Fig.~\ref{fig:k_scaling}(b)): as $K$ increases, the $\mathrm{FD}_{r,6}$ of 8- and 4-step generation continues to drop substantially, narrowing the gap to 50-step generation.

This effect is not confined to the base model at $\lambda=0$. Figure~\ref{fig:k_lambda_scaling} shows how Precision and Recall change with the auxiliary-conditioning scale $\lambda$ for various $K$ values. With $K=1$, both metrics remain nearly flat as $\lambda$ varies.
As $K$ increases, the response to $\lambda$ becomes progressively stronger: moving $\lambda$ away from zero shifts the same trained network from the higher-Precision base model toward higher Recall.

\iflatexml
\begin{figure}[t]
  \centering
  \includegraphics[width=\linewidth]{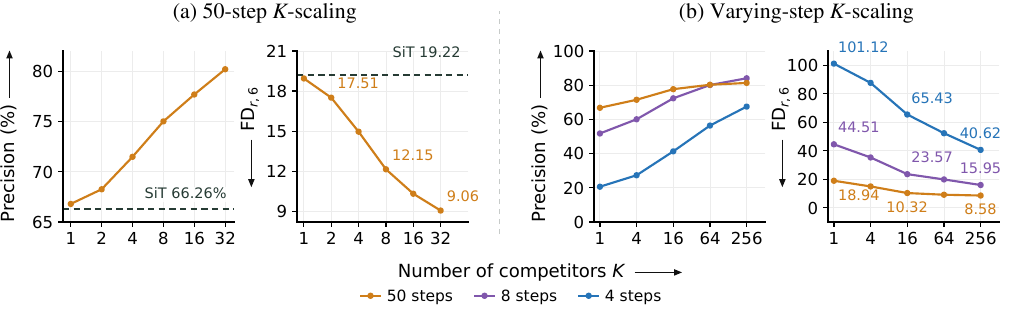}
\else
\par\medskip
\noindent\begin{minipage}{\linewidth}
  \captionsetup{type=figure}
  \centering
  
  \includegraphics[width=\linewidth]{cfm-figure6-k-scaling-panel-titles-unified-20261120.pdf}
\fi
  \caption{\textbf{Stronger competition increasingly reshapes the base model.}
(a) Under 50-step sampling, increasing the number of competitors $K$ steadily raises Precision while lowering $\mathrm{FD}_{r,6}$.
(b) The same trend is amplified under shorter sampling trajectories: 4- and 8-step generation continues to improve at larger $K$, progressively narrowing the gap to the 50-step reference.}
  \label{fig:k_scaling}
\iflatexml
\end{figure}
\else
\end{minipage}
\par\medskip
\fi

\iflatexml
\begin{figure}[t]
  \centering
  \includegraphics[width=\linewidth]{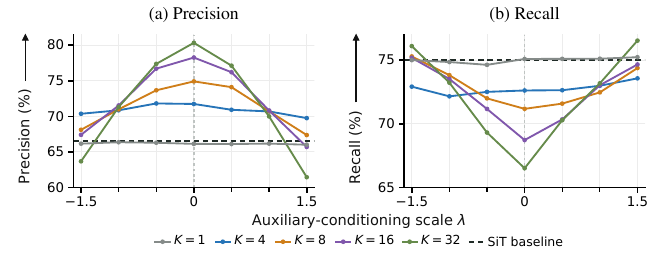}
\else
\par\medskip
\noindent\begin{minipage}{\linewidth}
  \centering
  \makebox[0.80\linewidth][l]{\hspace*{-0.03545455\linewidth}%
    \includegraphics[width=0.80\linewidth]{cfm-figure9-lambda-scaling-refined-20261120.pdf}}
  \captionsetup{type=figure}
\fi
  
  \caption{
\textbf{Competition reshapes the generator across the full range of auxiliary-conditioning scales.}
Precision (left) and Recall (right) as the auxiliary-conditioning scale
$\lambda$ varies for models post-trained with different numbers of competing
predictions $K$. The dashed horizontal line denotes pretrained SiT baseline. All results use $\gamma=0$.
}
  \label{fig:k_lambda_scaling}
\iflatexml
\end{figure}
\else
\end{minipage}
\par\medskip
\fi

\section{Conclusion}
A standard flow-matching model is trained toward a single operating point, leaving movement along the quality--coverage trade-off largely to inference-time guidance. In this work, we have shown that with WTA post-training, a single model instead holds two complementary behaviors: its base model operates at high fidelity, while auxiliary conditioning moves it toward high coverage. The auxiliary-conditioning scale $\lambda$ provides smooth control over this quality--coverage trade-off and can be used on its own or combined with existing guidance methods. Most notably, competition also changes how the generator responds to guidance. For pretrained models, stronger guidance improves quality at the cost of coverage; for WTA-post-trained models, it improves quality while largely preserving coverage, and at high-coverage operating points it can even increase Recall and within-class diversity. 
A practical limitation of our method is its additional training cost: while it does not require training models from scratch, selecting the winner involves evaluating $K$ candidate predictions for each training state, and this cost grows with $K$. 
Looking ahead, the auxiliary codes themselves merit closer study. Although WTA training only asks predictions to compete for each training target, without requiring any code to play a consistent role across states or sampling trajectories, a fixed code works well at inference: strengthening the auxiliary conditioning progressively increases coverage, and this combines well with guidance. The auxiliary-conditioned predictions thus appear to learn more structure than the objective explicitly asks for. Understanding this structure better, and learning or using auxiliary conditions in more adaptive ways, could further expand the quality--coverage frontier.

\ificlrpreprint
  \clearpage
\fi
\bibliographystyle{iclr2027_conference}
\bibliography{references}

\clearpage
\appendix
\section*{Supplementary Material}
\pdfbookmark[0]{Supplementary Material}{supplementary-material}
This supplement provides implementation and evaluation details, additional
quantitative results, inference ablations, analyses of generated variation,
and computational costs.

\subsection*{Contents}
\iflatexml
\else
\makeatletter
\begingroup
  \setcounter{tocdepth}{2}
  \@starttoc{apt}
\endgroup
\let\supp@addtocontents\addtocontents
\renewcommand{\addtocontents}[2]{%
  \def\supp@extension{#1}%
  \def\supp@toc{toc}%
  \ifx\supp@extension\supp@toc
    \supp@addtocontents{apt}{#2}%
  \else
    \supp@addtocontents{#1}{#2}%
  \fi
}
\makeatother
\clearpage
\fi

\section{Implementation and Evaluation Protocols}
\label{sec:supp-protocols}
\subsection{Backbone-Specific WTA Implementations}
\label{sec:supp-backbone-implementation}

We implement WTA by augmenting each pretrained generator with an auxiliary
code-conditioning pathway. This section specifies the architecture,
initialization, native regression loss, and post-training configuration
for SiT, JiT, RAE, and RAEv2.

\subsubsection{Shared training construction}
\label{sec:supp-backbone-common}
We use the notation of Eqs.~(\ref{eq:wta_codes})--(\ref{eq:wta_objective}):
$x_0$ denotes a clean training sample in the generator's input space,
$\epsilon$ is the Gaussian noise used to construct $x_t$, and
$z\sim\mathcal N(0,I_d)$ is an independent auxiliary code, with $d=8$.
The pretrained backbone already conditions on the timestep $t$ and class
label $y$, represented by embeddings $e_t(t)$ and $e_y(y)$.
Depending on the backbone, these embeddings provide conditioning through
its existing modulation modules, through in-context tokens processed
together with image tokens, or through both.

We denote this complete auxiliary-conditioning pathway by $c_\phi$.
The pathway first maps the auxiliary code to an embedding through $E_\phi$.
This additional conditioning embedding is then incorporated through the backbone's native
conditioning mechanism.
For modulation-based injection, the auxiliary contribution is combined
with the native conditioning along the modulation pathway.
For context-token injection, the auxiliary embedding is added to the
corresponding native context tokens.
The backbone-specific descriptions below specify the exact injection points.
All added parameters are collectively denoted by $\phi$.
We load the pretrained generator weights and initialize the added
pathway to contribute zero to the forward computation.
Both the shared generator parameters $\theta$ and auxiliary parameters $\phi$
are updated by this objective.

\subsubsection{SiT}
\label{sec:supp-backbone-sit}
\paragraph{Representation and native conditioning.}
SiT-XL/2~\citep{ma2024sit} operates on $4\times32\times32$ VAE latents
for $256\times256$ images. We retain the pretrained VAE, sample its
posterior during training, and apply the native latent scale $0.18215$.
The generator has 28 Transformer blocks of width $H=1152$, 16 attention
heads, and latent patch size 2. Each Transformer block uses native AdaLN
modulation to condition on $e_t(t)+e_y(y)$. Its modulation mapping
$M_{\mathrm{native}}^{(\ell)}$, implemented as SiLU followed by a linear
layer, produces $6H$ values: a shift, scale, and residual gate for each
of the attention and MLP sublayers. Shifts and scales modulate LayerNorm outputs.
Residual gates scale the sublayer outputs. These vectors are broadcast
over spatial tokens.

\paragraph{Auxiliary conditioning.}
The code embedding $E_\phi$ is a two-layer SiLU MLP shared across blocks,
with dimensions $8\to1152\to1152$:
\begin{equation}
 E_\phi(z)=W_2\operatorname{SiLU}(W_1z+b_1)+b_2.
 \label{eq:supp-backbone-code-mlp}
\end{equation}
Following the native AdaLN modulation structure, we introduce an
additional auxiliary modulation mapping
$M_\phi^{(\ell)}:\mathbb R^H\to\mathbb R^{6H}$ in each block, also
implemented as SiLU followed by a linear layer. It maps the auxiliary
embedding to additional shifts, scales, and residual gates, which are
added to their native counterparts:
\begin{equation}
 M_{\mathrm{native}}^{(\ell)}\!\left(e_t(t)+e_y(y)\right)
 +c_\phi^{(\ell)}(z),\qquad
 c_\phi^{(\ell)}(z)=M_\phi^{(\ell)}\!\left(E_\phi(z)\right).
 \label{eq:supp-backbone-adaln}
\end{equation}
The final output layer similarly receives an added $2H$-dimensional
shift and scale from its own SiLU--Linear projection of $E_\phi(z)$.

At initialization, the first embedding weight $W_1$ has independent Gaussian entries with
standard deviation $0.02$. The parameters $b_1$, $W_2$, and $b_2$ are zero.
All added modulation projections use Xavier-uniform weights and zero
biases. Thus the auxiliary contribution is initially zero, while the
nonzero projection weights provide a gradient path to the embedding's
final layer.

\subsubsection{JiT}
\label{sec:supp-backbone-jit}
\paragraph{Representation and native conditioning.}
JiT-H/16~\citep{li2025jit} predicts clean images in pixel space. It has
32 blocks of width $H=1280$, 16 attention heads, $16\times16$ image
patches, and a patch-embedding bottleneck of width 256. Each Transformer
block uses native AdaLN modulation to condition on $e_t(t)+e_y(y)$.
JiT also uses native in-context conditioning through 32 class-context tokens:
with
native context-position embeddings $p_j$, these are
$h_j^{\mathrm{ctx}}=e_y(y)+p_j$ for $j=1,\ldots,32$.
They are concatenated with the image tokens immediately before the
eleventh Transformer block and participate in subsequent self-attention.

\paragraph{Auxiliary conditioning.}
We construct the auxiliary embedding $E_\phi(z)$ using a two-layer SiLU MLP
with dimensions $8\to1280\to1280$, and set
\begin{equation}
 c_\phi(z)=E_\phi(z).
\end{equation}
Following these two native conditioning mechanisms, we add the same
auxiliary embedding to the input of the native AdaLN modulation modules
and to every class-context token:
\begin{equation}
 e_t(t)+e_y(y)+c_\phi(z),\qquad
 h_j^{\mathrm{ctx}}\leftarrow h_j^{\mathrm{ctx}}+c_\phi(z).
 \label{eq:supp-backbone-jit-injection}
\end{equation}
The JiT model in Table~\ref{tab:shared_predictor} uses a variant that
additionally adds this same auxiliary embedding to the image patch tokens
before the Transformer blocks.

The auxiliary pathway is initialized to contribute zero, preserving the
pretrained predictor at the start of post-training.

\subsubsection{RAE}
\label{sec:supp-backbone-raev1}
\paragraph{Representation and native conditioning.}
We retain the pretrained representation autoencoder~\citep{zheng2026diffusion},
using a DINOv2-with-registers Base encoder with $224\times224$ encoder
input, a ViT-XL image decoder, and the native latent normalization
statistics. The stage-two generator operates on $768\times16\times16$
latents. Its DDT consists of 28 backbone blocks of width $H_b=1152$
and two wide-head blocks of width $H_h=2048$, with patch size 1 and
16 attention heads in each stack. These stacks belong to the generative
Transformer. The stage-one image encoder and decoder remain frozen.
Both stacks use native RMSNorm-based modulation. Their modulation
mappings, implemented as SiLU followed by a linear layer, produce shifts,
scales, and residual gates for the attention and MLP sublayers.
Shifts and scales modulate RMSNorm outputs. Residual gates scale the
sublayer outputs.
The backbone modulation input is $\operatorname{SiLU}(e_t(t)+e_y(y))$.
For an encoded image-token feature $h_i$, the wide-head modulation input
is the native projection of $\operatorname{SiLU}(e_t(t)+h_i)$ to width
$H_h$. This input therefore also depends on the encoded image features.

\paragraph{Auxiliary conditioning.}
Two independent two-layer SiLU MLPs embed the same code:
$E_\phi^{b}:8\to1152\to1152$ and
$E_\phi^{h}:8\to2048\to2048$.
Following the native modulation structure, we introduce an auxiliary
modulation mapping $M_\phi^{(\ell)}:\mathbb R^{H_s}\to\mathbb R^{6H_s}$
for each block in stack $s\in\{b,h\}$. Implemented as SiLU followed by a
linear layer, it produces additional shifts, scales, and residual gates:
\begin{equation}
 c_\phi^{(\ell)}(z)=M_\phi^{(\ell)}\!\left(E_\phi^{s}(z)\right),
 \label{eq:supp-backbone-raev1-injection}
\end{equation}
which is added to that block's native modulation output. The final layer
receives an auxiliary shift and scale from a separate $2H_h$-dimensional
SiLU--Linear projection of $E_\phi^{h}(z)$.
Both embeddings and all added projections follow the SiT initialization:
Gaussian first-layer embedding weights with standard deviation $0.02$,
zero final-layer embedding weights, zero embedding biases, and
Xavier-uniform projection weights with zero biases.

\FloatBarrier
\subsubsection{RAEv2}
\label{sec:supp-backbone-raev2}
\paragraph{Representation and native conditioning.}
RAEv2 uses the pretrained DINOv3-L representation with features from
layers $11,13,15,17,19,21,23$, a ViT-XL image decoder, and the native
normalization statistics. The stage-two input has shape
$1024\times16\times16$. The generator has 28 backbone blocks of width
1440 with 20 attention heads, followed by two wide-head blocks of width
2048 with 16 heads. The backbone uses native in-context conditioning:
it processes image tokens together with four native timestep tokens and
eight native class-context tokens, denoted by $h_j^{\mathrm{ctx}}$.

\paragraph{Auxiliary conditioning.}
The code embedding $E_\phi$ maps $z$ to additive conditioning for the eight
native class-context tokens, each of width 1440. Here $c_\phi=E_\phi$, and
$[c_\phi(z)]_j$ denotes the auxiliary contribution at context position $j$.
Following this native context-token mechanism, we add each contribution to
the corresponding native class-context token:
\begin{equation}
 h_j^{\mathrm{ctx}}\leftarrow h_j^{\mathrm{ctx}}+[c_\phi(z)]_j,
 \qquad j=1,\ldots,8.
 \label{eq:supp-backbone-raev2-injection}
\end{equation}
The auxiliary pathway is initialized to contribute zero, preserving the
pretrained predictor at the start of post-training.

\subsubsection{Native losses and post-training settings}
\label{sec:supp-backbone-training}
We post-train all four generators on the ImageNet-1K training set at
$256\times256$ image resolution.
We retain each backbone's native image preprocessing and representation
normalization.

\paragraph{Regression targets and time sampling.}
We retain each backbone's native prediction parameterization,
interpolation convention, time-sampling scheme, and regression loss.
WTA uses this regression loss to score auxiliary-code candidates and
select the winner. For RAEv2, the candidate loss is the sum of the native
main-head and base-head velocity losses, with equal weights.

\paragraph{Post-training configuration.}
Table~\ref{tab:supp-backbone-training} summarizes the post-training
configurations. We follow each backbone's original training configuration
for the global batch size, defined as the total number of training images
per optimizer update across devices and gradient-accumulation steps.
All four models use auxiliary-code dimension $d=8$, zero weight decay,
and class dropout probability $0.1$.
SiT uses AdamW with moment coefficients $(0.9,0.999)$ and a constant
learning rate. JiT uses AdamW with $(0.9,0.95)$ and linearly warms up the
learning rate over one epoch before holding it constant. RAE uses AdamW with
$(0.9,0.95)$ and a constant learning rate. RAEv2 retains GMuon, with momentum
$0.95$ and Nesterov updates, for eligible hidden weight matrices and AdamW
with $(0.9,0.95)$ for embeddings, biases, normalization parameters, output
layers, and the auxiliary embedding. The RAEv2 main learning rate is held for
the first half epoch and then decays linearly to $10^{-5}$ at epoch ten. The
auxiliary embedding has a $20\times$ learning-rate multiplier and its own
500-update linear warmup, after which it follows the decay of the main
schedule.

\begin{table}[htbp]
 \centering
 \caption{WTA post-training configurations for the four backbones.}
 \label{tab:supp-backbone-training}
 \small
 \begin{tabular}{@{}lrrrr@{}}
 \toprule
 Setting & SiT & JiT & RAE & RAEv2 \\
 \midrule
 Global batch size & 256 & 1024 & 1024 & 1024 \\
 Post-training epochs & 10 & 10 & 10 & 10 \\
 Main learning rate & $10^{-4}$ & $2\!\times\!10^{-4}$ & $2\!\times\!10^{-4}$ & $5\!\times\!10^{-5}$ \\
 EMA decay & 0.9999 & 0.9999 & 0.9995 & 0.9995 \\
 Gradient-norm limit & --- & --- & 1 & 1 \\
 Random seed & 0 & 0 & 42 & 42 \\
 \bottomrule
 \end{tabular}
\end{table}

\subsubsection{Inference and evaluation protocols}
\label{sec:supp-native-evaluation}
For the base-model comparison in Table~\ref{tab:shared_predictor}, we follow
the sampling protocol used in each model's original paper. Each pretrained
generator and its $K=8$ post-trained counterpart use the same sampling
settings, with the post-trained model evaluated at $\lambda=0$.
Each entry uses 50K generated images. The original models' inference-time
guidance is disabled. The sampling settings are:
\begin{itemize}
  \item \textbf{SiT:} Euler SDE sampling with 250 steps, a linear diffusion
  schedule with diffusion norm 1, and a final mean step of size 0.04.
  \item \textbf{JiT:} Heun ODE sampling with 50 steps (99 model evaluations).
  \item \textbf{RAE:} The native Euler ODE configuration with the step-count
  argument set to 50 and time shift $\sqrt{48}$. The implementation constructs
  a grid of 50 time points.
  \item \textbf{RAEv2:} Euler ODE sampling with 50 steps and time shift 8.
  The model's internal guidance scale is set to 1, recovering its unguided prediction.
\end{itemize}
The SiT entries in Table~\ref{tab:shared_predictor} use 250-step SDE sampling,
whereas Table~\ref{tab:guided_performance} uses the controlled 50-step Euler ODE
recipe and reports guidance-selected results.

\paragraph{Metric evaluation.}
We compute FID, Inception Score, Precision, and Recall using the ADM
evaluation suite~\citep{dhariwal2021adm}.
We measure within-class perceptual variation using the original DreamSim
ensemble~\citep{fu2023dreamsim} and cosine distance between normalized embeddings.
We average this distance over all unordered image pairs within each class,
and then average the class means with equal class weights.
We compute $\mathrm{FD}_{r,6}$ using the released FD-Loss implementation and
ImageNet reference statistics of \citet{yang2026fdloss}.
Let $d_j$ be the unnormalized Fr\'echet distance in representation $j$ and
$a_j$ its released training--validation reference distance. We report
\begin{equation}
  \mathrm{FD}_{r,6}=\frac{1}{6}\sum_{j=1}^{6}\frac{d_j}{a_j}.
  \label{eq:supp-fdr6-evaluation}
\end{equation}
The six representations are Inception (ADM), ConvNeXt, DINOv2, MAE,
SigLIP, and CLIP, with respective normalization constants
$1.68$, $56.87$, $14.19$, $0.04$, $0.60$, and $5.60$.
We use the released global-feature variants (the \texttt{cls} entries for
DINOv2, MAE, SigLIP, and CLIP), rather than patch-average variants.
Normalization is applied only when forming $\mathrm{FD}_{r,6}$: reported
individual FD components retain their raw values. We use the FD-Loss
implementation for all six $\mathrm{FD}_{r,6}$ components and the ADM evaluation
suite for standalone FID.

\FloatBarrier
\subsection{Implementation Details of the One-Dimensional Toy}
\label{sec:supp-toy-implementation}

\paragraph{Source and target distributions.}
The toy in Fig.~\ref{fig:toy} transports a standard Gaussian source
$\epsilon\sim\mathcal{N}(0,1)$ to a mixture of two compactly supported modes.
Let $B\sim\operatorname{Beta}(3,3)$ and let $M$, independently of $B$, equal
$-0.75$ with probability $0.25$ and $+0.75$ with probability $0.75$.
We generate target samples as $x_0=M+0.6(2B-1)$.
Equivalently, the target density is
\begin{equation}
 p_{\mathrm{data}}(x)=0.25\,q_{0.6}(x+0.75)+0.75\,q_{0.6}(x-0.75),
 \qquad
 q_w(s)=\frac{15}{16w}\left(1-\frac{s^2}{w^2}\right)^2
 \mathbf{1}_{\{|s|\leq w\}}.
 \label{eq:supp-toy-target}
\end{equation}
The two supports are $[-1.35,-0.15]$ and $[0.15,1.35]$, with zero target
density in the intervening gap. Training draws fresh source and target
samples independently at each update.

\paragraph{Velocity predictor and auxiliary conditioning.}
The scalar velocity predictor uses three fully connected hidden layers of
width $128$, each followed by SiLU, and a linear scalar output layer.
Its input is the $18$-dimensional vector
$[x_t,t,\{\sin(j\pi t),\cos(j\pi t)\}_{j=1}^{8}]$.
An auxiliary code $z\sim\mathcal{N}(0,I_8)$ passes through a separate
$8\!\to\!128\!\to\!128$ MLP with SiLU after both linear layers.
For each predictor hidden layer, a separate linear map projects this code
representation to $256$ values, split into a scale $s_\ell(z)$ and a shift
$b_\ell(z)$ of dimension $128$ each. If $a_\ell$ is the hidden layer's
pre-activation, conditioning gives
\begin{equation}
 h_\ell=\operatorname{SiLU}\!\left(
 a_\ell\odot[1+\lambda s_\ell(z)]+\lambda b_\ell(z)\right).
 \label{eq:supp-toy-conditioning}
\end{equation}
All scale-and-shift projection weights and biases are initialized to zero,
so enabling the auxiliary pathway initially preserves the pretrained
velocity field.

\paragraph{Flow-matching pretraining.}
We sample $t\sim\mathcal{U}[0,1]$ and form
$x_t=(1-t)\epsilon+tx_0$, with velocity target $u=x_0-\epsilon$.
The predictor is trained with the mean squared error
$\mathbb{E}[(v_\theta(x_t,t)-u)^2]$, with auxiliary conditioning disabled.
We use Adam with batch size $1{,}024$, default moment coefficients
$(0.9,0.999)$, and gradient-norm clipping at $10$.
The learning rate is $2\times10^{-4}$ for the first $20{,}000$ updates,
$5\times10^{-5}$ for the next $10{,}000$, $10^{-5}$ for the next $10{,}000$,
and $5\times10^{-6}$ thereafter.
We train for $50{,}000$ updates and monitor convergence using held-out error
against the conditional-mean
velocity field computed by numerical quadrature, together with the
Wasserstein-1 distance between generated samples and the target distribution. We use the $45{,}000$-update checkpoint, selected based on held-out velocity
error, as both the FM baseline and the initialization for WTA post-training.

\paragraph{WTA post-training.}
For each freshly sampled $(x_t,t,u)$, we draw $K=8$ independent codes
$z_k\sim\mathcal{N}(0,I_8)$ and compute eight predictions using the shared
predictor with $\lambda=1$.
We minimize the batch average of
$\min_{k\in\{1,\ldots,8\}}(v_\theta(x_t,t;c_\phi(z_k))-u)^2$.
Only the minimizing candidate contributes gradients for each training
example; both the shared predictor and auxiliary conditioning parameters
are updated. Post-training uses a fresh Adam optimizer for $40{,}000$
updates, with learning rate $5\times10^{-5}$ for the first $20{,}000$
updates and $10^{-5}$ for the remaining $20{,}000$.
The batch size, Adam moment coefficients, and gradient clipping are the
same as in FM pretraining.

\paragraph{Inference.}
We generate $8{,}192$ samples per setting from standard Gaussian initial
states, reusing the same initial states across FM, auxiliary-on, and
auxiliary-off inference. Given the simplicity of this one-dimensional toy,
we use a correspondingly short four-step sampling process to highlight
differences in off-support generation and mode coverage, illustrating the
intuition behind our method. Sampling uses four explicit Euler updates from
$t=0$ to $t=1$, with $t_i=i/4$ and step size $1/4$:
\begin{equation}
 x^{(i+1)}=x^{(i)}+\tfrac14 v_\theta(x^{(i)},t_i;c_\phi(z)),
 \qquad i=0,1,2,3.
 \label{eq:supp-toy-euler}
\end{equation}
Auxiliary-on inference uses $\lambda=1$ and draws one independent Gaussian
code per trajectory, held fixed across all four updates. Auxiliary-off
inference uses the same post-trained predictor with $\lambda=0$, bypassing
the code-conditioning pathway. Standard FM sampling uses the pretrained
predictor with conditioning disabled.
For the trajectory illustration in Fig.~\ref{fig:toy}(c), we use two fixed
auxiliary codes at $\lambda=1$.

\FloatBarrier
\section{Efficient Guidance Search with Matched Evaluation Budgets}
\label{sec:efficient-guidance-search}

\subsection{Guidance Search and Evaluation}
\label{sec:supp-guidance-search}

\paragraph{Constructing guidance curves.}
We construct guidance curves to compare the overall quality--coverage trends of all methods in the teaser. We write guided predictions as $v+\gamma(v-v_G)$, where $v$ is the generator prediction ($v_\lambda$ for our method). For CFG, $v_G$ is the unconditional prediction of the same model. For AG, $v_G$ is the prediction of a weaker guiding model. Within each guidance mechanism, our method and the compared methods use the same grid with spacing $0.25$, covering $\gamma\in[0,4]$ for CFG and $\gamma\in[0,3]$ for AG.

\begin{figure}[!htbp]
  \centering
  \includegraphics[width=\linewidth]{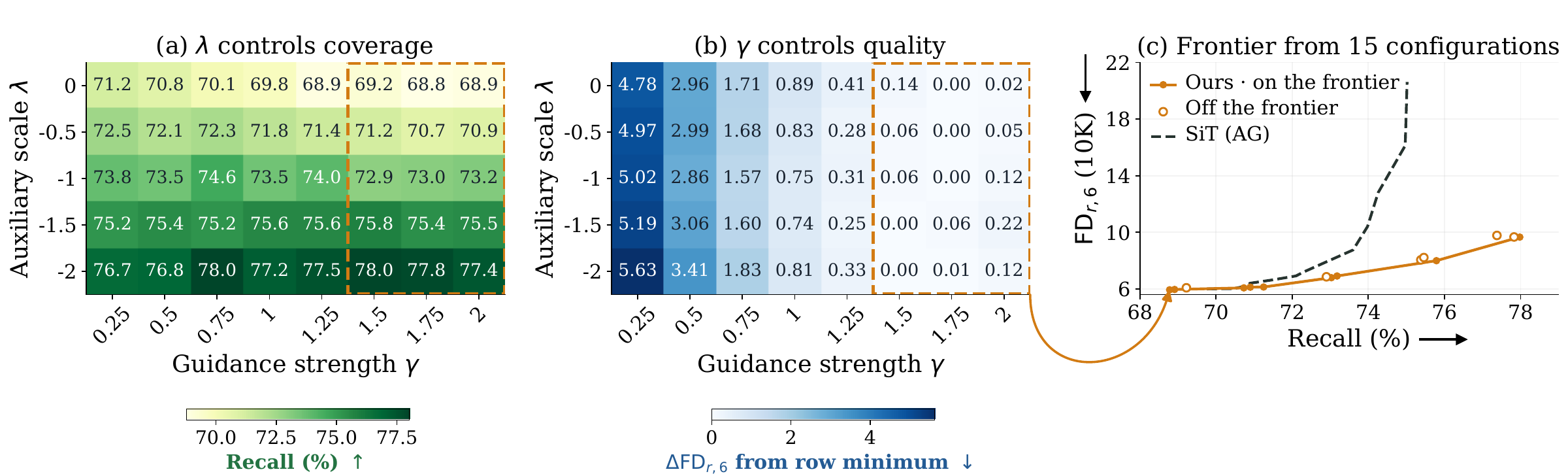}
  \caption{Quality and coverage across $\lambda$ and $\gamma$. (a,b) Heatmaps over the same combinations of auxiliary-conditioning scale $\lambda$ (rows) and guidance strength $\gamma$ (columns). Colors show (a) Recall and (b) $\mathrm{FD}_{r,6}$ above the minimum measured within each row. (c) Frontier from the 15 orange-boxed configurations, compared with the Pareto frontier of 19 evaluated SiT AutoGuidance configurations. Filled and hollow markers indicate nondominated and dominated configurations, respectively. All evaluations use 10K samples and 50-step Euler sampling.}
  \label{fig:search-budget-guidance-controls}
\end{figure}

\paragraph{Our frontier.}
While our method introduces two inference-time controls, the auxiliary-conditioning scale $\lambda$ and guidance strength $\gamma$, an exhaustive search over the two-dimensional space is unnecessary, as the two controls exhibit consistent and complementary effects. As shown in \cref{fig:search-budget-guidance-controls}(a,b), varying $\lambda$ primarily shifts the accessible Recall range, while $\gamma$ improves quality within that range. Notably, favorable guidance strengths remain similar across $\lambda$, around $\gamma\in[1.5,2.0]$. This enables a structured search that first identifies a favorable guidance interval and then varies $\lambda$ to progressively expand coverage. A compact $5\times3$ grid using only 15 configurations already recovers the main quality--coverage frontier gains, as shown in \cref{fig:search-budget-guidance-controls}(c).

\FloatBarrier

We further compare the two guidance searches under the same evaluation budget. For SiT with AutoGuidance, we evaluate 19 guidance configurations using a 13-point coarse sweep over $\gamma\in[0,3]$ with spacing $0.25$, followed by six local refinements. For our method, five coarse evaluations at $\lambda=0$ first locate the favorable guidance range, and two additional evaluations refine it to $\gamma\in\{1.5,1.75,2.0\}$. We then evaluate these three guidance strengths at each of four additional auxiliary scales, for the same total of 19 configurations. As shown in \cref{fig:search-budget-guidance-controls}(c), the resulting compact $5\times3$ grid achieves a substantially stronger quality--coverage frontier under the matched evaluation budget. Thus, the gain does not require an exhaustive two-dimensional sweep. It follows from a small structured search that exploits the complementary roles of $\lambda$ and $\gamma$.

\begingroup
\begin{table}[!htbp]
  \centering
  \caption{\textbf{Representative 10K operating points for our guided frontier.}
  Our $K=8$ model with 50-step Euler sampling.}
  \label{tab:supp-sit-operating-10k}
  \footnotesize
  \begin{tabular*}{0.72\linewidth}{@{\extracolsep{\fill}}cccc@{}}
    \toprule
    $\lambda$ & $\gamma$ & Recall (\%) $\uparrow$ & $\mathrm{FD}_{r,6}$ $\downarrow$ \\
    \midrule
    $-0.50$ & $1.75$ & $70.73$ & $6.071$ \\
    $-1.50$ & $1.50$ & $75.79$ & $7.992$ \\
    $-1.75$ & $1.50$ & $77.04$ & $8.788$ \\
    $-2.00$ & $1.50$ & $77.98$ & $9.658$ \\
    \bottomrule
  \end{tabular*}
\end{table}
\endgroup

\paragraph{Table results.}
Table~\ref{tab:guided_performance} reports the best performance of each method in two settings: without a Recall constraint and at a high-Recall target of approximately $67\%$. For all methods, we select the configuration with the lowest $\mathrm{FD}_{r,6}$ in each setting, choosing $\gamma$ from the ranges above on a grid with spacing $0.05$. We perform this search coarse-to-fine rather than evaluating every grid point. A coarse sweep at spacing $0.25$, with 10K generated samples per configuration, identifies the best-performing $\gamma$ regions, which we then search at spacing $0.05$. We evaluate the selected configurations with 50K images and 50 Euler updates, and Table~\ref{tab:supp-guidance-settings} lists their settings.

\begingroup
\begin{table}[!htbp]
  \centering
  \caption{\textbf{Guidance settings for Table~\ref{tab:guided_performance}.}
  Entries report $\gamma$, or $(\lambda,\gamma)$ for our $K=8$ model.
  For CFG, the conventional scale is $s=1+\gamma$.
  Best settings minimize $\mathrm{FD}_{r,6}$ among evaluated 50K configurations.
  A dagger denotes a below-target reference.}
  \label{tab:supp-guidance-settings}
  \footnotesize
  \begin{tabular}{@{}llcc@{}}
    \toprule
    Model & Guidance & Best $\mathrm{FD}_{r,6}$ & Recall $\approx67\%$ \\
    \midrule
    CAFM & CFG & $0.90$ & $0.05$ \\
    REPA & CFG & $1.05$ & $0.05$ \\
    MixFlow & CFG & $1.60$ & $0.00^{\dagger}$ \\
    Self-Flow & CFG & $0.95$ & $0.00^{\dagger}$ \\
    StableVM & CFG & $1.00$ & $0.00$ \\
    DeepFlow & CFG & $1.10$ & $0.00$ \\
    iREPA & CFG & $1.10$ & $0.00^{\dagger}$ \\
    SiT & CFG & $1.45$ & $0.00^{\dagger}$ \\
    SiT & AG & $2.35$ & $0.00^{\dagger}$ \\
    Ours ($K=8$) & AG & $(-0.05,1.80)$ & $(-1.25,1.50)$ \\
    \bottomrule
  \end{tabular}
\end{table}
\endgroup

\FloatBarrier

\FloatBarrier
\section{Robustness Across Evaluation Seeds}
\label{sec:eval-seed-robustness}

To assess the stability of our reported quality--coverage improvements,
we evaluate six representative operating points from our main-paper
quality--coverage frontier, together with the best-$\mathrm{FD}_{r,6}$
operating points of competing methods, across three independent evaluation
seeds using 50K samples per seed. The trained checkpoints and inference
configurations are fixed, while initial noise, class labels, and auxiliary
codes (when applicable) vary across seeds.

As shown in \cref{tab:eval-seed-robustness}, the results remain stable
across evaluation seeds, with sample standard deviations at most
0.0164 in $\mathrm{FD}_{r,6}$ and 0.0051 in Recall.
At the high-coverage operating point, we obtain
$\mathrm{FD}_{r,6}=5.9033\pm0.0079$ at
$0.6784\pm0.0034$ Recall. The best-$\mathrm{FD}_{r,6}$ operating
points of the competing methods are similarly stable across evaluation seeds.
These results demonstrate low evaluation-seed variability across both our
representative frontier points and the measured competing-method operating
points.

\begin{table}[H]
    \centering
    \caption{
        \textbf{Robustness across evaluation seeds.}
        Mean $\pm$ sample standard deviation across three independent
        evaluation seeds, with 50K samples and 50-step Euler sampling
        per evaluation. Training checkpoints and inference settings
        are fixed. We report representative frontier points for our method
        and the best-$\mathrm{FD}_{r,6}$ operating point for each competing
        method. Reported variability reflects evaluation seeds, not training
        seeds or confidence intervals.
    }
    \label{tab:eval-seed-robustness}
    \small
    \begin{tabular}{llcc}
        \toprule
        Method / point & Inference setting
        & $\mathrm{FD}_{r,6}\downarrow$
        & Recall ([0,1]) $\uparrow$ \\
        \midrule
        Ours (A) & $(\lambda,\gamma)=(-0.05,1.80)$ & $4.4544\pm0.0132$ & $0.6064\pm0.0021$ \\
        Ours (B) & $(\lambda,\gamma)=(-0.50,1.75)$ & $4.5886\pm0.0115$ & $0.6303\pm0.0051$ \\
        Ours (C) & $(\lambda,\gamma)=(-0.60,1.15)$ & $5.1757\pm0.0083$ & $0.6437\pm0.0023$ \\
        Ours (D) & $(\lambda,\gamma)=(-1.00,1.25)$ & $5.5910\pm0.0121$ & $0.6652\pm0.0044$ \\
        Ours (E) & $(\lambda,\gamma)=(-1.25,1.50)$ & $5.9033\pm0.0079$ & $0.6784\pm0.0034$ \\
        Ours (F) & $(\lambda,\gamma)=(-0.70,1.00)$ & $5.6073\pm0.0137$ & $0.6522\pm0.0043$ \\
        \midrule
        SiT & AG $2.35$ (best) & $4.5306\pm0.0068$ & $0.6208\pm0.0012$ \\
        CAFM & CFG $1.90$ (best) & $4.8562\pm0.0164$ & $0.5722\pm0.0011$ \\
        \bottomrule
    \end{tabular}
\end{table}

\clearpage

\section{Additional Evaluation and Ablations}
\label{sec:supp-quantitative}
\label{sec:supp-ablations}
\subsection{Precision--Recall Evaluation}
\label{sec:supp-precision-recall}

\begin{figure}[!htbp]
  \centering
  \includegraphics[width=0.62\linewidth]{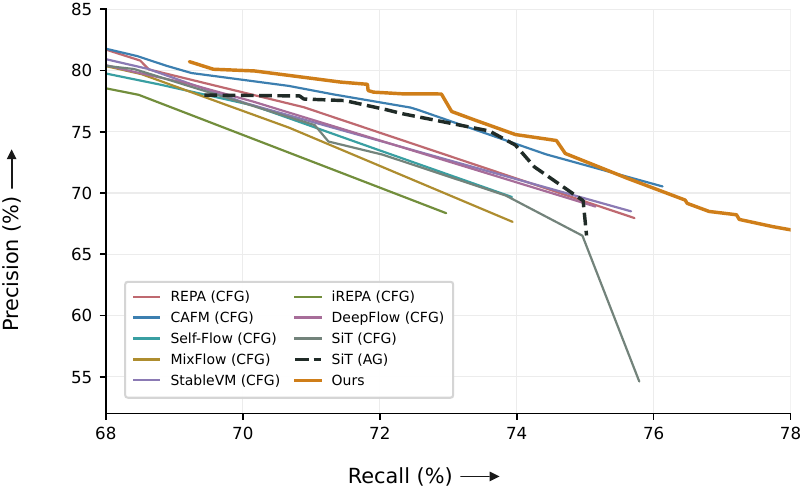}
  \caption{\textbf{Precision--Recall comparison in the high-coverage regime.}
  Results on ImageNet-256 using 10K generated samples per operating point
  and a 50-step Euler ODE solver.
  Curves summarize guidance sweeps; WTA also varies the auxiliary-conditioning
  scale. CFG and AG denote classifier-free guidance and AutoGuidance, respectively.
  Higher Precision and Recall are better.}
  \label{fig:supp-precision-recall}
\end{figure}

We also report Precision--Recall curves on ImageNet-256, comparing WTA with
recent flow-matching methods across inference-time guidance settings.
Precision and Recall characterize sample fidelity and distribution coverage,
respectively~\citep{kynkaanniemi2019improved}.
Each operating point is evaluated using 10K generated samples.
Figure~\ref{fig:supp-precision-recall} provides an additional view of the
fidelity--coverage trade-off: WTA maintains competitive Precision while
extending to higher Recall, highlighting its advantage in high-coverage generation.

\FloatBarrier
\subsection{Additional Model Capacity and Continued Training}
\label{sec:supp-capacity-training}

\subsubsection{Are the Gains Due to Additional Model Capacity?}
\label{sec:supp-capacity}

The $K=1$ control retains the same auxiliary-conditioning architecture and
parameter count as our $K=8$ model, but is post-trained without winner
competition. Both models share the pretrained initialization and are compared
at matched post-training exposure. This allows us to test whether the gains
can be explained by the added model capacity alone.

At $\gamma=0$, we have already shown that the $K=1$ model remains close to
pretrained SiT at $\lambda=0$ and responds little to changes in $\lambda$
(Figs.~\ref{fig:k_scaling} and~\ref{fig:k_lambda_scaling}). We further
evaluate this capacity-matched control by varying both $\lambda$ and $\gamma$.

\begin{figure}[htbp]
  \centering
  \includegraphics[width=0.72\linewidth]{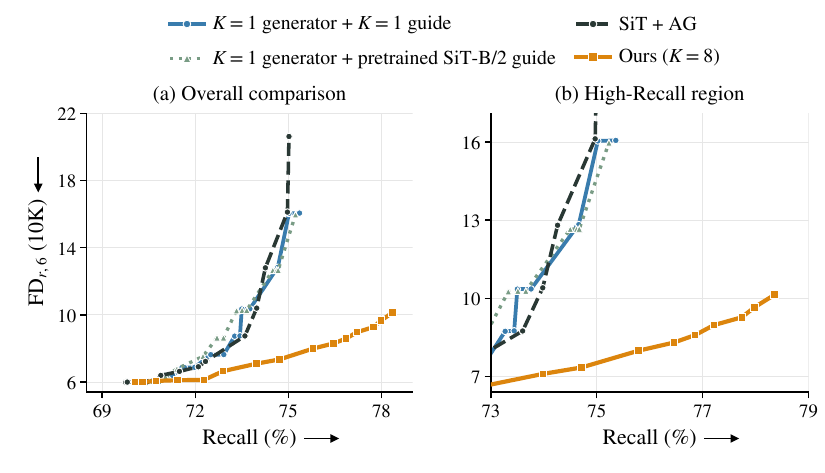}
  \caption{\textbf{Additional parameters without competition do not reproduce the
  high-Recall WTA frontier.} Capacity-matched $K=1$ controls compared with pretrained SiT+AG and our $K=8$ model. The $K=1$ generator is paired with either a $K=1$ post-trained or an original pretrained SiT-B/2 guide. (a) Overall comparison. (b) High-Recall region. All results use 10K samples and 50-step Euler sampling.}
  \label{fig:supp-capacity-frontier}
\end{figure}

Following the main AutoGuidance setup, we use the $K=1$ post-trained SiT-XL/2
as the generator and a SiT-B/2 model also post-trained with $K=1$ as the guide.
Figure~\ref{fig:supp-capacity-frontier} shows that its guided frontier remains
close to the pretrained SiT+AG baseline at high Recall and does not reproduce
the high-coverage improvement of our $K=8$ model.

The generator uses a Gaussian auxiliary code fixed along each trajectory.
We sweep its auxiliary-conditioning scale
$\lambda\in\{-1.5,-1,-0.5,0,0.5,1,1.5\}$ and AutoGuidance strength
$\gamma\in\{0,0.25,\ldots,2\}$, including the native $\lambda=1$ setting.

\paragraph{Additional check with original SiT-B.}
We also pair the $K=1$ post-trained generator with original pretrained SiT-B
using the same sweep. This configuration likewise remains close to pretrained
SiT+AG at high Recall (Fig.~\ref{fig:supp-capacity-frontier}). Together with
the results at $\gamma=0$, these controls indicate that added capacity alone
does not explain the observed high-coverage improvement.

\FloatBarrier
\subsubsection{Continued Standard Flow-Matching Training}
\label{sec:supp-sit-fm-continuation}

The $K=1$ experiment retains the auxiliary architecture while removing
competition. We complement it with standard-FM continuation controls,
which test additional training without the competitive objective.
Table~\ref{tab:supp-sit-fm-continuation} reports SiT-XL/2
continuation results under 10K, 50-step Euler evaluation with $\gamma=0$.
FID remains close to the pretrained value after 50 additional
training epochs. These controls therefore do not reproduce the change seen
with competitive post-training.

\begin{table}[htbp]
  \centering
  \caption{\textbf{Continued standard FM changes SiT-XL/2 only modestly.}
  10K samples, 50-step Euler sampling, $\gamma=0$.}
  \label{tab:supp-sit-fm-continuation}
  \begin{tabular}{lrr}
    \toprule
    Model & FID $\downarrow$ & IS $\uparrow$ \\
    \midrule
    Pretrained SiT-XL/2 & 14.41 & 118.60 \\
    FM continuation, 50 epochs & 14.31 & 120.90 \\
    \bottomrule
  \end{tabular}
\end{table}

\paragraph{Larger training batches.}
We continue standard FM training of SiT-B/2 for 20 epochs at global batch
sizes of 256, 1024, and 8192, starting from the same pretrained initialization
(Table~\ref{tab:supp-sit-fm-batch}).
Across batch sizes, the overall changes relative to the pretrained model
remain limited. Increasing the batch size does not produce a consistent
improvement from standard-FM continuation.

\begin{table}[htbp]
  \centering
  \caption{\textbf{Training batch size in standard-FM continuation.}
  SiT-B/2 after 20 additional data epochs, evaluated using 10K samples, 50-step Euler sampling, and $\gamma=0$.
  The pretrained row is the common initialization before continuation.}
  \label{tab:supp-sit-fm-batch}
  \begin{tabular}{lrr}
    \toprule
    Model / training batch size & FID $\downarrow$ & IS $\uparrow$ \\
    \midrule
    Pretrained SiT-B/2 & 33.40 & 51.42 \\
    FM continuation, batch 256 & 32.84 & 53.45 \\
    FM continuation, batch 1024 & 31.76 & 56.18 \\
    FM continuation, batch 8192 & 32.81 & 53.20 \\
    \bottomrule
  \end{tabular}
\end{table}

\FloatBarrier
\subsection{Signs of Auxiliary Conditioning}
\label{sec:supp-sit-lambda-sign}

The main-text analysis shows similar overall responses to positive and negative
auxiliary-conditioning scales, with negative scales generally yielding lower
$\mathrm{FD}_{r,6}$. We therefore use negative $\lambda$ values for the guided
frontier presented in the main paper. Here, we additionally construct the
frontier using positive $\lambda$ values.

\begin{figure}[!htbp]
  \centering
  \includegraphics[width=0.56\linewidth]{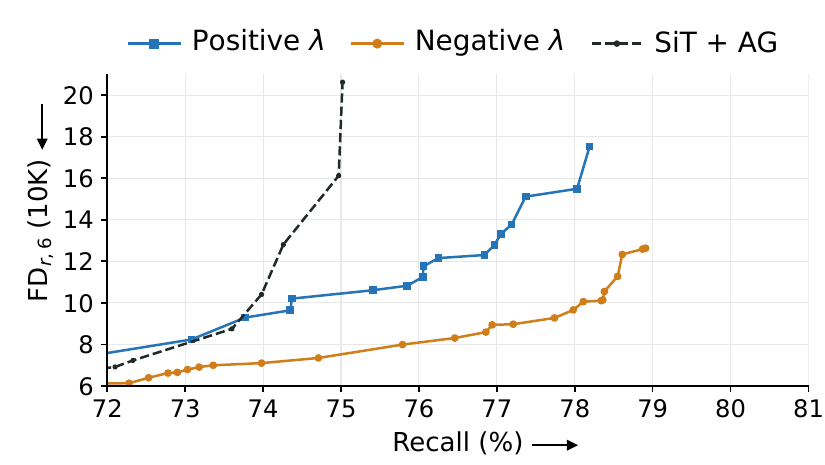}
  \caption{\textbf{Both signs of auxiliary conditioning improve the high-coverage frontier.}
  $K=8$ guided frontiers for positive and negative $\lambda$, compared
  with pretrained SiT AutoGuidance baseline. Both signs improve the frontier in the
  high-coverage region.
  All results use 10K samples and 50-step Euler sampling.}
  \label{fig:supp-sit-lambda-sign}
\end{figure}

Figure~\ref{fig:supp-sit-lambda-sign} shows that both signs improve the
quality--coverage frontier over pretrained SiT with AutoGuidance in the
high-coverage region. Empirically, negative scales generally provide a slightly
better trade-off. 

\FloatBarrier
\subsection{Effect of Competition Level \texorpdfstring{$K$}{K}}
\label{sec:supp-sit-competition}

\subsubsection{SiT at $\gamma=0$}
\label{sec:supp-sit-lambda-response}

\begin{figure}[!htbp]
  \centering
  \includegraphics[width=0.90\linewidth]{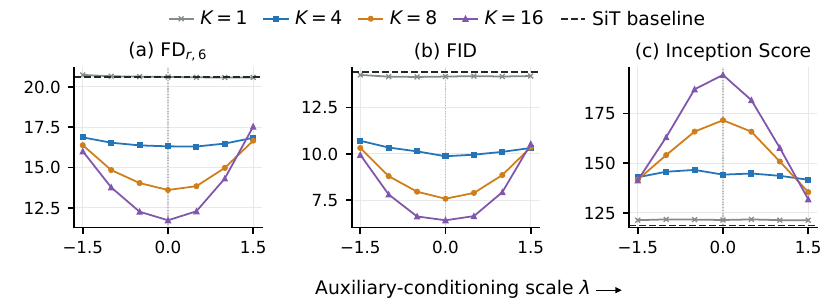}
  \caption{\textbf{Competition changes the response to auxiliary conditioning.}
  (a) $\mathrm{FD}_{r,6}$, (b) FID, and (c) Inception Score for
  $K\in\{1,4,8,16\}$ with $\gamma=0$.
  The dashed horizontal lines denote pretrained SiT.
  All curves use the same $\lambda$ values and 10K samples with
  50-step Euler sampling.}
  \label{fig:supp-sit-lambda-response}
\end{figure}

Figure~\ref{fig:supp-sit-lambda-response} complements
Fig.~\ref{fig:k_lambda_scaling} with $\mathrm{FD}_{r,6}$, FID, and Inception
Score as the auxiliary-conditioning scale $\lambda$ varies with $\gamma=0$.
We use the same SiT-XL/2 checkpoint for each $K$
throughout these comparisons, with $K\in\{1,4,8,16\}$.

The $K=1$ curves remain nearly flat and close to pretrained SiT. With
competition, the post-trained base models at $\lambda=0$ attain lower
$\mathrm{FD}_{r,6}$ and FID, and higher Inception Score. Varying $\lambda$
exposes a broader response for $K=8$ and $K=16$ than for $K=4$.
Together with the Precision--Recall curves, these results show that
competition changes both the base model and the behaviors accessible through
auxiliary conditioning.

\FloatBarrier
\subsubsection{Guided Frontiers on SiT}
\label{sec:supp-sit-k-frontiers}

We next combine auxiliary conditioning with AutoGuidance. For each fixed
$\lambda$, varying the inference-time guidance strength $\gamma$ traces a
quality--coverage trajectory.
Each generator is paired with a SiT-B/2 guiding model post-trained with the
same $K$.

\begin{figure}[!htbp]
  \centering
  \includegraphics[width=0.56\linewidth]{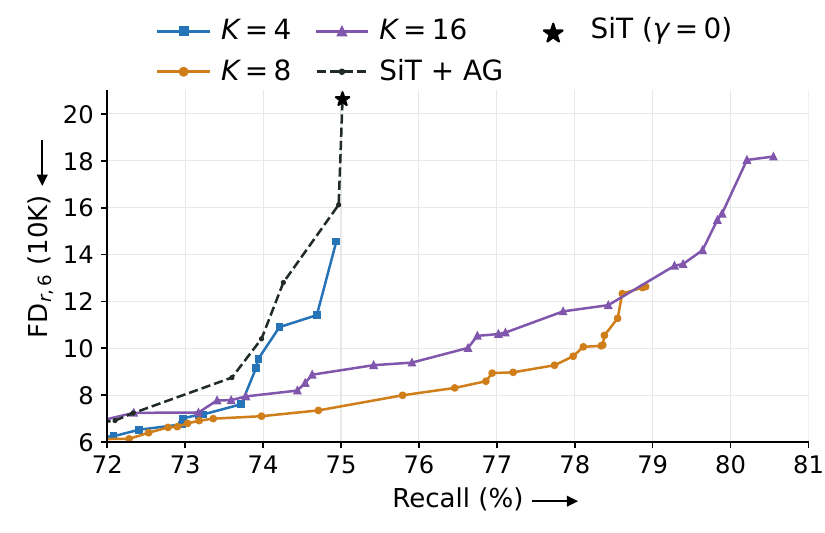}
  \caption{\textbf{The high-coverage frontier improvement persists across competition levels.}
  Measured guided frontiers for $K=4,8,16$, with pretrained SiT
  AutoGuidance and its $\gamma=0$ point as references.
  The plot shows the high-coverage region.
  All results use 10K samples and
  50-step Euler sampling.}
  \label{fig:supp-sit-k-frontiers}
\end{figure}

Figure~\ref{fig:supp-sit-k-frontiers} shows that guided-frontier improvements
persist across $K=4,8,16$. Compared with pretrained SiT+AG, all three improve
the quality--coverage trade-off in the overlapping high-coverage region,
with larger $K$ extending the attained Recall range.

\FloatBarrier
\subsubsection{Guided Frontiers on RAEv2}
\label{sec:supp-raev2-competition}

\begin{figure}[!htbp]
  \centering
  \includegraphics[width=0.52\linewidth]{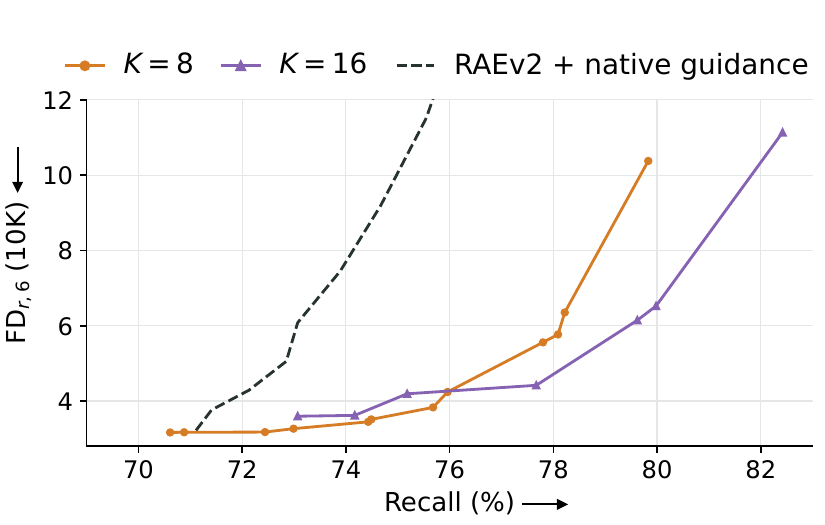}
  \caption{\textbf{The RAEv2 frontier improvement persists across competition levels.}
  Measured $\mathrm{FD}_{r,6}$--Recall frontiers for $K=8,16$ and pretrained
  RAEv2 with its native guidance. Each setting uses 10K samples
  and native Euler-50 sampling with timestep shift 8.
  Two pretrained-reference points have $\mathrm{FD}_{r,6}$ values above the plotted range.}
  \label{fig:supp-raev2-k-frontiers}
\end{figure}

We extend the $K=8$ comparison in Figure~\ref{fig:raev2_transfer} by including $K=16$.
For both competition levels, we sweep the auxiliary-conditioning scale $\lambda$
and the RAEv2-guidance strength $\gamma$, using RAEv2's native sampler
(Appendix~\ref{sec:supp-native-evaluation}) and guidance window $t\in[0.1,1.0]$.
Both competition levels improve the quality--coverage frontier over pretrained
RAEv2, and $K=16$ extends it further into the high-coverage region
(Figure~\ref{fig:supp-raev2-k-frontiers}).

\FloatBarrier
\subsection{Auxiliary-Code Sampling Policies}
\label{sec:supp-auxiliary-code-sampling}

We compare the policies introduced in \cref{sec:inference_guidance}:
\emph{fixed-code sampling} holds one Gaussian auxiliary code throughout each
trajectory, while \emph{resampled-code sampling} draws a new code at every
Euler update. Both use the same SiT-XL/2 WTA model ($K=8$),
with 10K images per setting, 50 Euler updates, and matched initial noises and
class labels. At $\lambda=0$, both recover the same post-trained base model.

\begin{figure}[!ht]
  \centering
  \includegraphics[width=\linewidth]{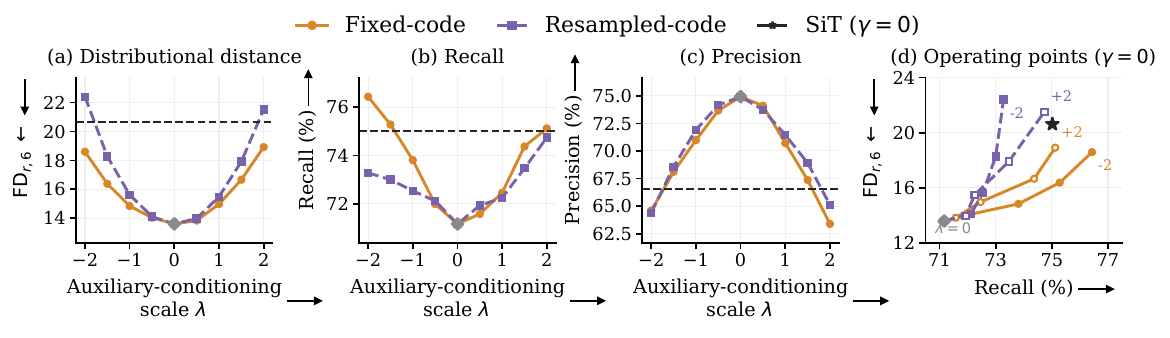}
  \caption{\textbf{Varying auxiliary conditioning under two sampling policies ($\gamma=0$).} SiT-XL/2 WTA ($K=8$),
  10K images per setting, 50 Euler updates. Panels (a--c) show
  $\mathrm{FD}_{r,6}$, Recall, and Precision versus auxiliary-conditioning scale;
  dashed black horizontal lines show pretrained SiT at $\gamma=0$. Panel (d) shows
  the same observations in Recall--$\mathrm{FD}_{r,6}$ space, with a black
  star for SiT. Gray diamonds denote the shared post-trained base model ($\lambda=0$).
  In (d), filled and open policy markers denote negative and positive
  scales, respectively; endpoint labels give $\lambda$.}
  \label{fig:supp-code-sampling-unguided}
\end{figure}

\paragraph{Varying the auxiliary-conditioning scale ($\gamma=0$).}
\Cref{fig:supp-code-sampling-unguided} shows similar overall responses to
auxiliary-conditioning strength under both sampling policies, with fixed-code
sampling generally yielding lower $\mathrm{FD}_{r,6}$ than resampled-code sampling.

\FloatBarrier
\FloatBarrier

\begin{figure}[!ht]
  \centering
  \includegraphics[width=0.95\linewidth]{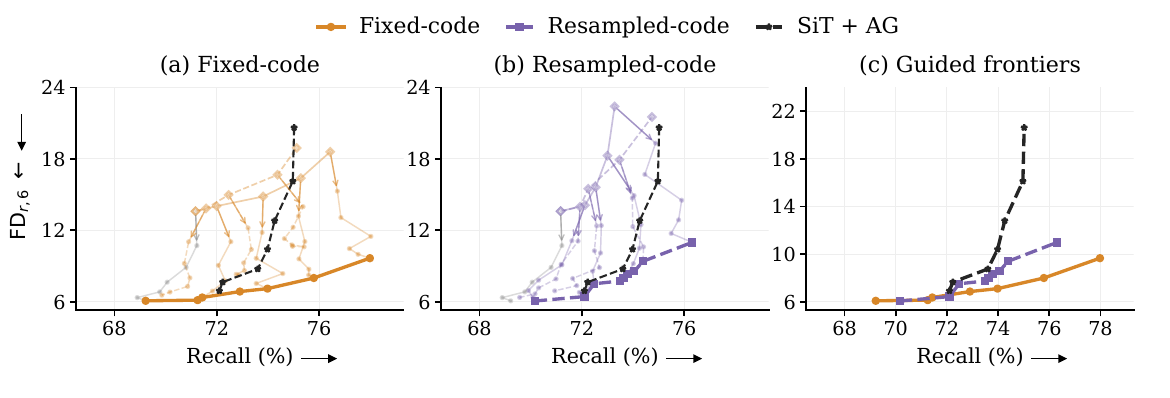}
  \caption{\textbf{Both code policies improve the guided frontier, with
  a larger broad-coverage gain from fixed codes.} All results use 10K samples and 50-step Euler sampling.
  In (a,b), pale diamonds connect the $\gamma=0$ operating points; pale paths vary
  $\gamma$ at fixed $\lambda$, with arrows pointing toward increasing
  guidance. Solid and dashed pale paths denote nonpositive and positive
  $\lambda$, respectively; gray denotes the post-trained base model ($\lambda=0$). (c) compares these
  frontiers directly with original SiT+AG, using the same axis ranges.
}
  \label{fig:supp-code-sampling-guided}
\end{figure}
\FloatBarrier

\paragraph{Guided frontiers.}
We compare both sampling policies using the same $(\lambda,\gamma)$ settings.
As shown in \cref{fig:supp-code-sampling-guided}, resampled-code sampling also
improves the quality--coverage frontier over SiT+AG, while fixed-code sampling
yields a stronger frontier, particularly at high recall.

\FloatBarrier

\paragraph{Interpretation.}
Holding the auxiliary code fixed may allow its influence to persist across
sampling steps, whereas resampling changes that influence at each step.
This offers a possible explanation for the stronger high-coverage frontier
observed with fixed-code sampling.

\FloatBarrier
\subsection{Reusable Auxiliary-Code Signatures}
\label{sec:supp-reusable-code-signatures}

To further investigate the effect of holding an auxiliary code fixed, we
analyze how individual codes shape generated samples across initial noises
and classes.

The experiments in this section use fixed-code sampling with $\lambda=1$, the training-time auxiliary-conditioning scale, and $\gamma=0$. 

\begin{figure}[!htbp]
  \centering
  \begin{subfigure}{\linewidth}
    \centering
    \includegraphics[width=\linewidth]{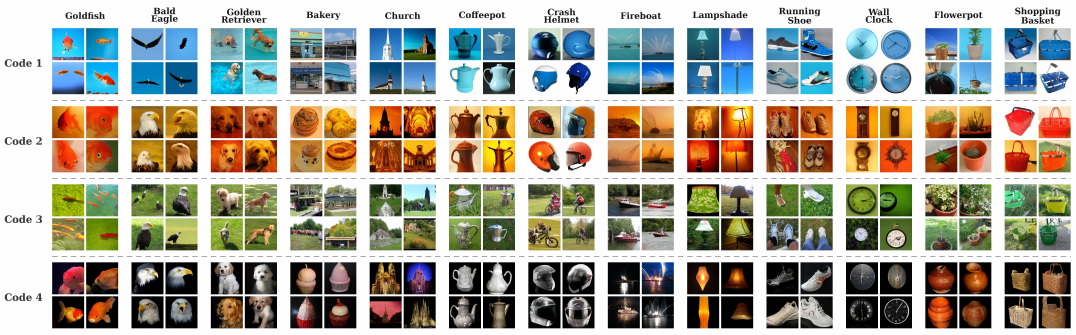}
    \caption{Reusable visual signatures across ImageNet classes.}
    \label{fig:supp-reusable-code-visuals-main}
  \end{subfigure}
  \par\medskip
  \begin{subfigure}{\linewidth}
    \centering
    \includegraphics[width=\linewidth]{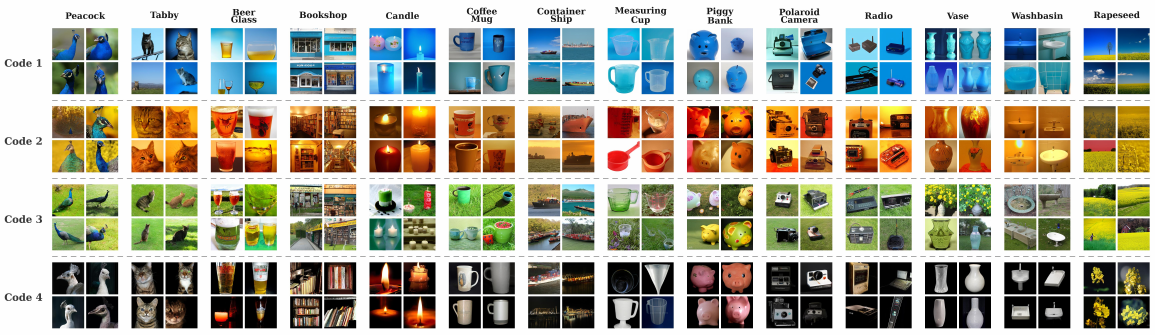}
    \caption{Additional classes with the same auxiliary codes.}
    \label{fig:supp-reusable-code-visuals-additional}
  \end{subfigure}
  \caption{\textbf{Samples generated with the same auxiliary code share recurring
  appearance patterns across source noises and classes.} Each row fixes one
  auxiliary code $z$ and each column one ImageNet class. The four
  samples at each row--column intersection vary only the initial noise
  $\epsilon$. Both panels use the same SiT-XL/2 WTA checkpoint,
  the same four auxiliary codes, and 50-step ODE sampling at
  $\lambda=1$, $\gamma=0$, holding the code fixed throughout each trajectory.
  Repeated tendencies in color, background, and composition appear both within
  and across classes.}
  \label{fig:supp-reusable-code-visuals}
\end{figure}

Although auxiliary codes are sampled independently during training and are not
assigned persistent roles, samples generated with the same fixed code exhibit
recurring appearance patterns across initial noises and classes
(\cref{fig:supp-reusable-code-visuals}).

\FloatBarrier

\iflatexml\else\Needspace{8\baselineskip}\fi
\subsubsection{Code Signatures Strengthen with Competition}
\label{sec:supp-reusable-code-competition}

We quantify how consistently these code-specific patterns recur across classes
and how this changes with $K$. Specifically, we test whether the auxiliary code
can be identified from generated samples on held-out classes. We fix 64
auxiliary codes for analysis and split 128 ImageNet classes into a reference
half and a held-out half.

\newpage
\noindent\makebox[\linewidth][c]{%
\begin{minipage}{0.88\linewidth}
  \begin{minipage}[c]{0.31\linewidth}
    \centering
    \includegraphics[width=\linewidth]{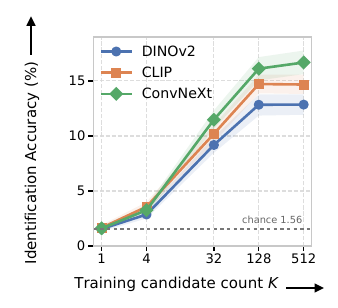}
  \end{minipage}\hfill
  \begin{minipage}[c]{0.64\linewidth}
    \captionsetup{type=figure,font=footnotesize,skip=0pt}
  \caption{\textbf{Code signatures strengthen with competition.}
  Identification accuracy on held-out classes, over 64 fixed analysis codes;
  chance is 1.56\%. The matched $K$ sweep uses 128 classes and four source-noise
  seeds, with 50-step ODE sampling and fixed auxiliary codes at
  $\lambda=1$, $\gamma=0$. Shaded bands are 95\% hierarchical class--seed bootstrap
  intervals.}
  \label{fig:supp-reusable-code-k-trend}
  \end{minipage}
\end{minipage}%
}
\par\medskip

We first define the change a code makes: for a given class and source noise,
we generate one sample with the code and one without, and compute the difference
of their features in a pretrained visual encoder's space (e.g., DINOv2).
We then subtract the average of this difference over all codes, leaving only
what is specific to this code. Averaging these code-specific differences over
the reference classes gives the code's signature. For each held-out sample,
we compute its code-specific difference in the same way, and assign it to the
code with the nearest signature. Chance is $1/64$, and identification succeeds
only if a code's effect recurs on classes never used to build its signature.

Across DINOv2, CLIP, and ConvNeXt features, identification accuracy is near
chance for the $K=1$ control and rises with competition over the measured range,
reaching roughly ten times chance at $K=512$
(\cref{fig:supp-reusable-code-k-trend}).

\FloatBarrier

\FloatBarrier

\subsection{Guidance Configurations}
\label{sec:supp-guidance-configurations}
\subsubsection{AutoGuidance Configuration}
\label{sec:supp-ag-design}

We compare alternative guiding models and auxiliary-conditioning settings.
The configuration used in our main experiments achieves the strongest
high-coverage frontier among those evaluated
(Fig.~\ref{fig:supp-ag-design-frontiers}).

\par\medskip
\noindent\begin{minipage}{\linewidth}
\captionsetup{type=figure}
  \centering
  \includegraphics[width=0.70\linewidth]{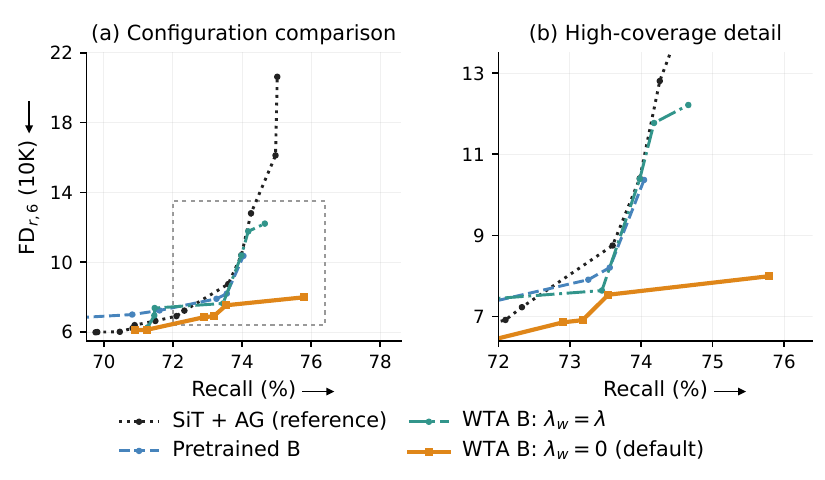}
  \caption{\textbf{AutoGuidance guiding-model configurations.}
  The three configurations share the same WTA-post-trained $K=8$ generator
  and the same $(\lambda,\gamma)$ settings; only the guiding model or its auxiliary
  conditioning changes. The default uses WTA-post-trained SiT-B/2 with
  $\lambda_w=0$.
  (a) Configuration comparison. (b) Detail of the boxed high-coverage region.
  SiT+AG uses the reference guidance sweep.
  All results use 10K samples and 50-step Euler sampling.}
  \label{fig:supp-ag-design-frontiers}
\end{minipage}
\par\medskip

\paragraph{Compared configurations.}
We hold the WTA-post-trained SiT-XL/2 generator fixed ($K=8$)
and compare three guiding-model configurations: the original pretrained
SiT-B/2; a WTA-post-trained SiT-B/2 ($K=8$) with auxiliary
conditioning enabled; and the same post-trained guiding model with auxiliary
conditioning disabled, as used in the main experiments.
For the enabled configuration, both models receive the same numerical
auxiliary code and conditioning scale, $\lambda_w=\lambda$.
For the default configuration, the guiding model uses $\lambda_w=0$.

\paragraph{Evaluation.}
All three configurations use the same generator conditioning scales
$\lambda\in\{-1.5,-1,-0.5,0.5,1,1.5\}$ and inference-time guidance strengths
$\gamma\in\{0.5,1,1.5,2\}$, following Eq.~\ref{eq:combined_guidance}.
Each setting uses 10K samples with matched initial noises and 50-step Euler
sampling. A Gaussian auxiliary code is held fixed along each sampling trajectory.
Pretrained SiT-XL/2 guided by pretrained SiT-B/2 provides the SiT+AG reference.

\subsubsection{Combining with Classifier-Free Guidance}
\label{sec:supp-cfg}
We also combine auxiliary conditioning with classifier-free guidance (CFG)
and compare our post-trained SiT-XL/2 model with $K=8$ against pretrained
SiT-XL/2 with CFG. The guiding prediction is the class-unconditional output
of the same model. The conditional and unconditional predictions use the same
auxiliary code and conditioning scale $\lambda$, with the code fixed along
each trajectory.
Figure~\ref{fig:supp-cfg-frontiers} shows that our post-trained model improves
the quality--coverage trade-off over pretrained SiT with CFG in the displayed
Recall range.

\par\medskip
\noindent\begin{minipage}{\linewidth}
\captionsetup{type=figure}
\centering
\includegraphics[width=0.56\linewidth]{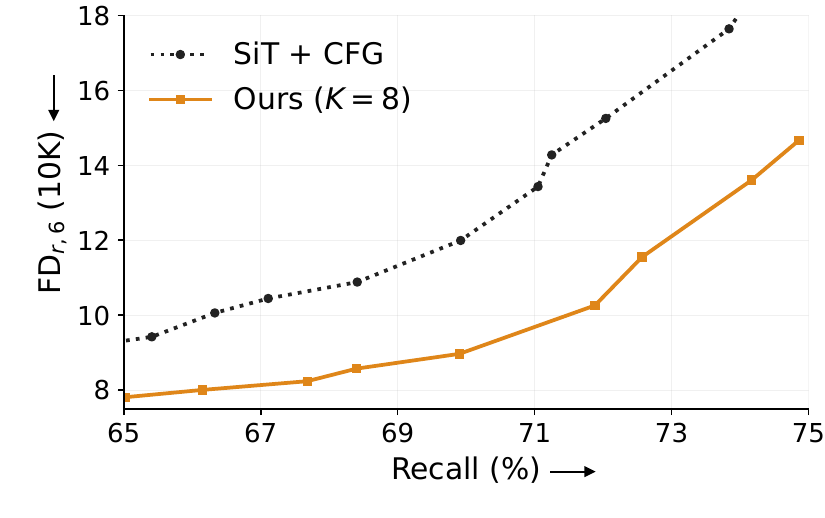}
\caption{Quality--coverage tradeoffs with classifier-free guidance.
Pretrained SiT+CFG is compared with our post-trained model using $K=8$.
Our curve is constructed from a joint $(\lambda,\gamma)$ sweep using
one fixed trained model.
All results use 10K images and 50-step Euler sampling.}
\label{fig:supp-cfg-frontiers}
\end{minipage}
\par\medskip

\clearpage

\section{Qualitative Results}
\label{sec:supp-variation}
\subsection{Qualitative Improvements in the Base Model}
\label{sec:supp-base-model-qualitative}

Our post-training improves the fidelity of the base model itself, even without auxiliary conditioning or additional guidance ($\lambda=0$, $\gamma=0$). The SiT and JiT examples in Figures~\ref{fig:supp-sit-qualitative} and~\ref{fig:supp-jit-qualitative} illustrate these improvements.

\begin{center}
\begin{minipage}{\textwidth}
\centering
\captionsetup{type=figure}
\includegraphics[width=\linewidth,height=0.74\textheight,keepaspectratio]{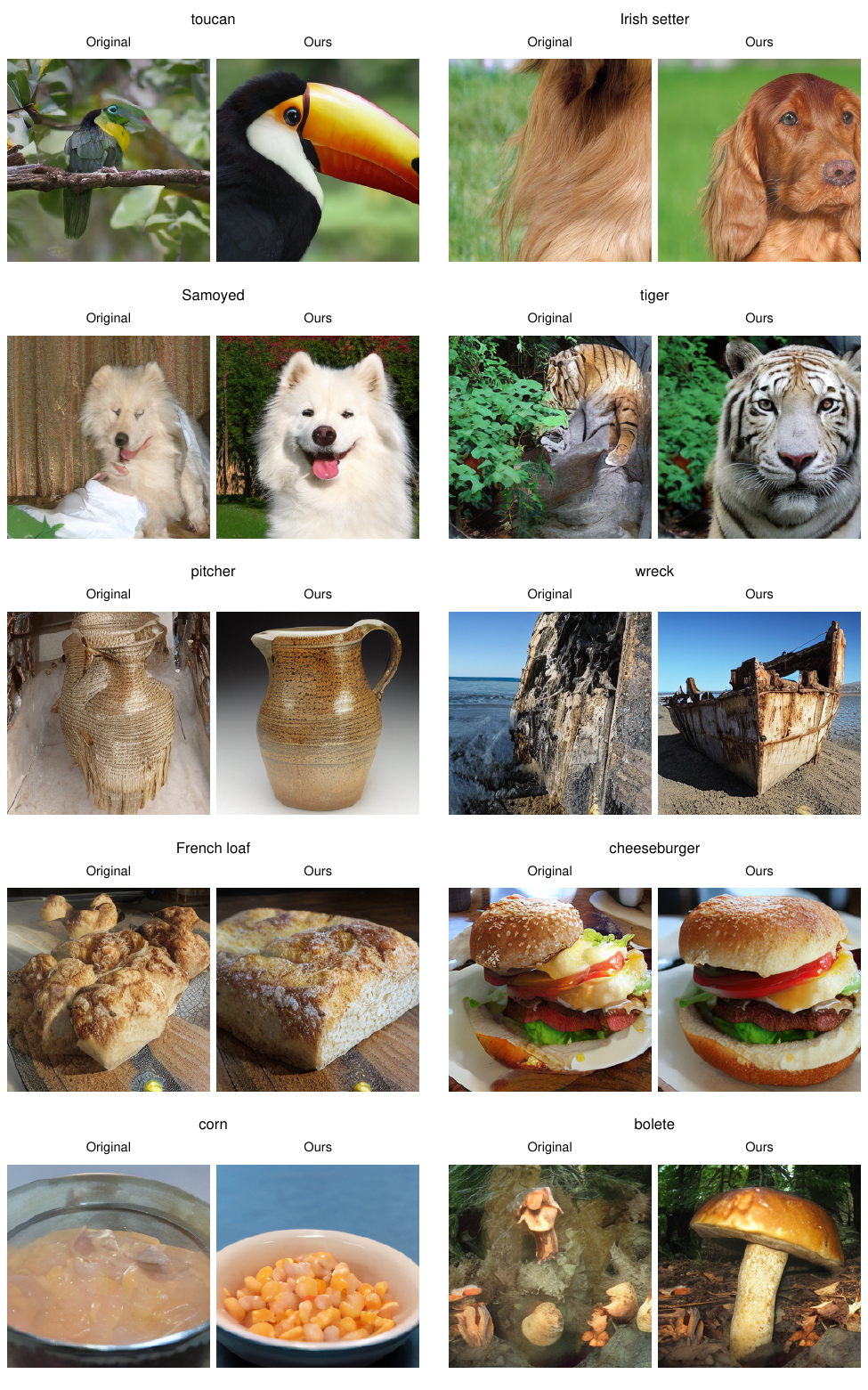}
\caption{\textbf{Base-model improvements on SiT-XL/2.} Ten pairs of ImageNet-256 samples from the original pretrained model (left) and our post-trained model (right). Each pair uses the same class and initial noise. Both models use 50 Euler steps without additional guidance. Our model uses $\lambda=0$ and $\gamma=0$.}
\label{fig:supp-sit-qualitative}
\end{minipage}
\end{center}
\clearpage

\begin{center}
\begin{minipage}{\textwidth}
\centering
\captionsetup{type=figure}
\includegraphics[width=\linewidth,height=0.74\textheight,keepaspectratio]{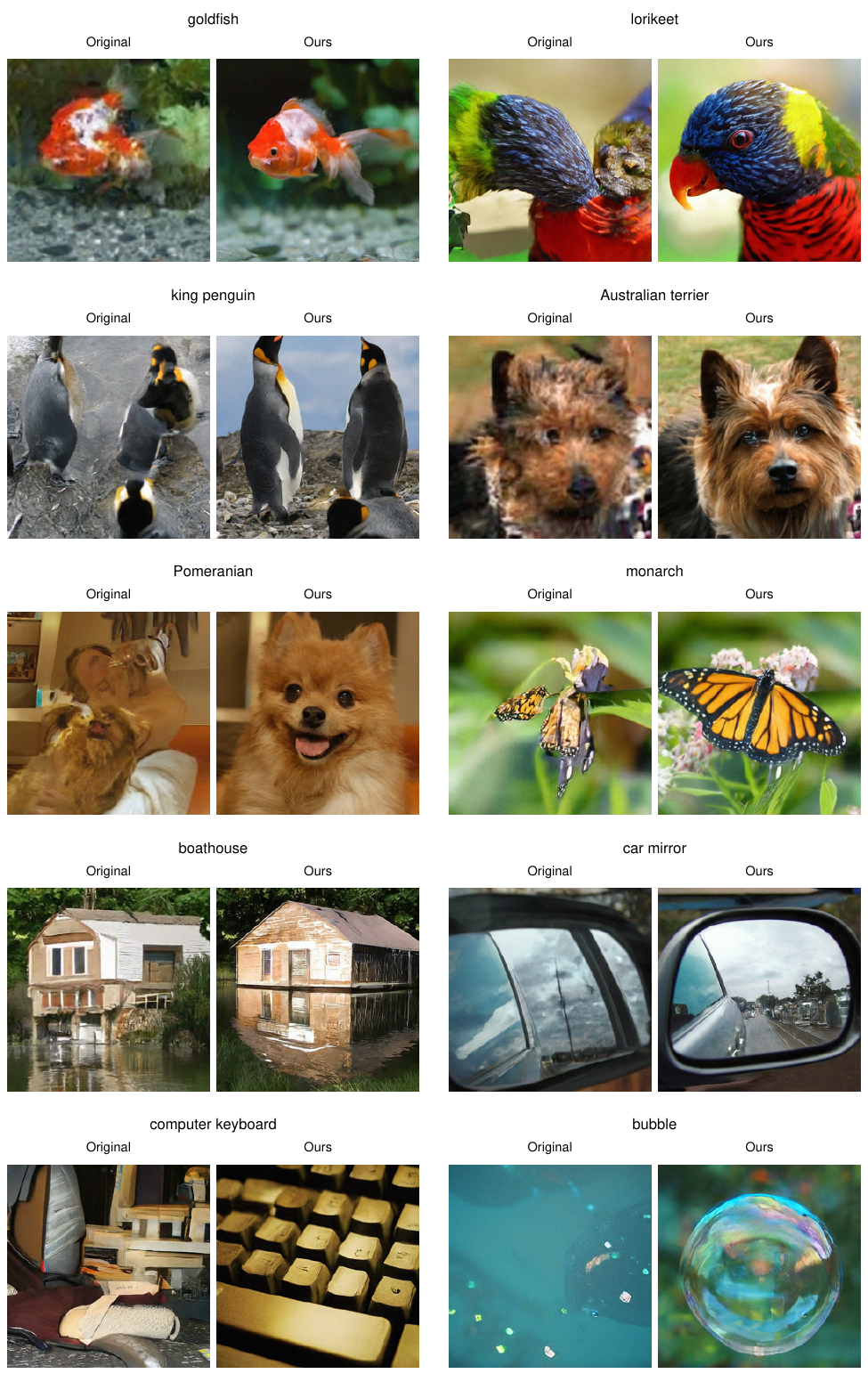}
\caption{\textbf{Base-model improvements on JiT-H/16.} Ten pairs of ImageNet-256 samples from the original pretrained model (left) and our post-trained model (right). Each pair uses the same class and initial noise. Both models use 50-step Heun sampling without additional guidance. Our model uses $\lambda=0$ and $\gamma=0$.}
\label{fig:supp-jit-qualitative}
\end{minipage}
\end{center}

\clearpage
\subsection{Structured Variation Under Auxiliary Conditioning and Guidance}
\label{sec:supp-code-strength-variation}
\label{sec:supp-lambda-sign-qualitative}

To examine how auxiliary conditioning and guidance change individual samples,
we visualize samples generated with different auxiliary codes across
auxiliary-conditioning scales and guidance strengths.

For each example, we fix the ImageNet class and initial noise and randomly
sample three auxiliary codes, each held fixed throughout sampling. Panel (a)
shows one code per row and $\lambda\in\{0,-0.5,-1,-1.5\}$ across columns at
$\gamma=0$. Panel (b) repeats this grid at AG strengths $\gamma=0,0.75,1.5$,
showing the shared $\lambda=0$ image once per block. All images are generated
by SiT-XL/2 with $K=8$ and 50-step Euler sampling.

At $\lambda=0$ and $\gamma=0$, sampling yields the post-trained base-model
image. As $|\lambda|$ increases, samples from different codes often share the
visual structure of this image while developing increasingly distinct
appearance and details. Increasing the AG strength $\gamma$ can further
accentuate these differences and sharpen image details.
\par\medskip
\noindent\begin{minipage}{\linewidth}
\centering
\input{figures/supplementary_code_strength_variation/01_volcano}
\end{minipage}\par
\clearpage
\noindent\begin{minipage}{\linewidth}
\centering
\input{figures/supplementary_code_strength_variation/02_lakeside}
\end{minipage}\par
\vspace{6pt}
\noindent\begin{minipage}{\linewidth}
\centering
\input{figures/supplementary_code_strength_variation/03_pizza}
\end{minipage}\par
\clearpage
\noindent\begin{minipage}{\linewidth}
\centering
\input{figures/supplementary_code_strength_variation/04_white_wolf}
\end{minipage}\par
\vspace{6pt}
\noindent\begin{minipage}{\linewidth}
\centering
\input{figures/supplementary_code_strength_variation/05_goldfish}
\end{minipage}\par
\clearpage
\noindent\begin{minipage}{\linewidth}
\centering
\input{figures/supplementary_code_strength_variation/06_jack_o_lantern}
\end{minipage}\par
\vspace{6pt}
\noindent\begin{minipage}{\linewidth}
\centering
\input{figures/supplementary_code_strength_variation/07_daisy}
\end{minipage}\par
\clearpage
\noindent\begin{minipage}{\linewidth}
\centering
\input{figures/supplementary_code_strength_variation/08_golden_retriever}
\end{minipage}\par
\vspace{6pt}
\noindent\begin{minipage}{\linewidth}
\centering
\input{figures/supplementary_code_strength_variation/09_cockatoo}
\end{minipage}\par
\clearpage

\FloatBarrier
We also extend these visualizations to both positive and negative
auxiliary-conditioning scales, complementing the quantitative comparison
in Appendix~\ref{sec:supp-sit-lambda-sign}.

\begin{figure}[!ht]
  \centering
  \includegraphics[width=\linewidth]{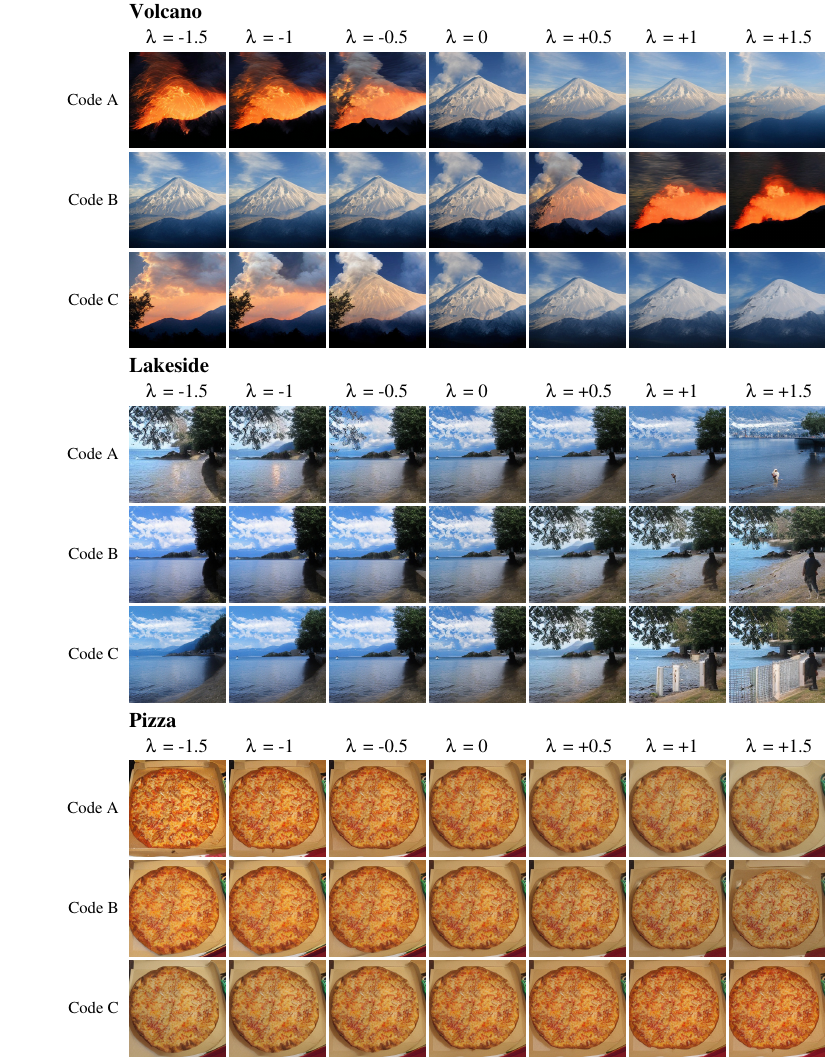}
  \caption{\textbf{Positive and negative auxiliary-conditioning scales.}
  Columns show $\lambda\in\{-1.5,-1,-0.5,0,0.5,1,1.5\}$.
  Within each class, the initial noise is fixed; within each row, the auxiliary
  code is fixed. All samples use $\gamma=0$, SiT-XL/2 $K=8$, and 50-step Euler sampling.}
  \label{fig:supp-lambda-sign-qualitative-1}
\end{figure}
\clearpage

\begin{figure}[!ht]
  \centering
  \includegraphics[width=\linewidth]{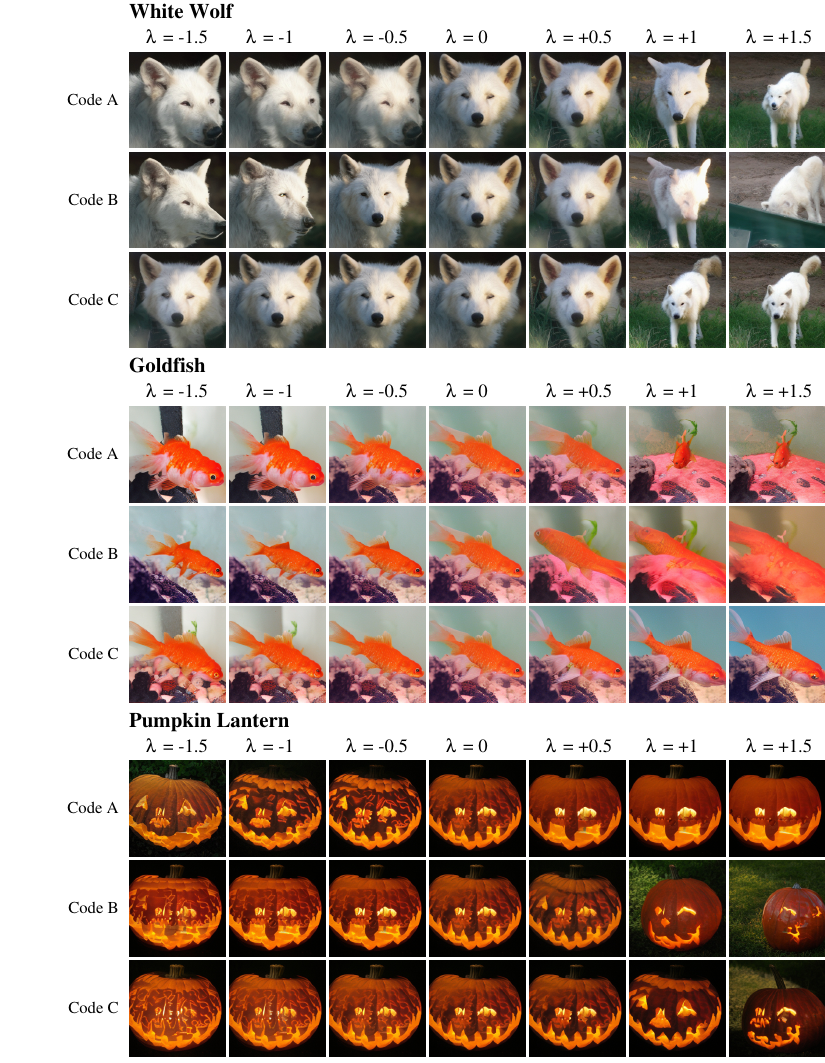}
  \caption{\textbf{Positive and negative auxiliary-conditioning scales on additional classes.}
  The layout and sampling settings follow Figure~\ref{fig:supp-lambda-sign-qualitative-1}.}
  \label{fig:supp-lambda-sign-qualitative-2}
\end{figure}
\clearpage

\begin{figure}[!ht]
  \centering
  \includegraphics[width=\linewidth]{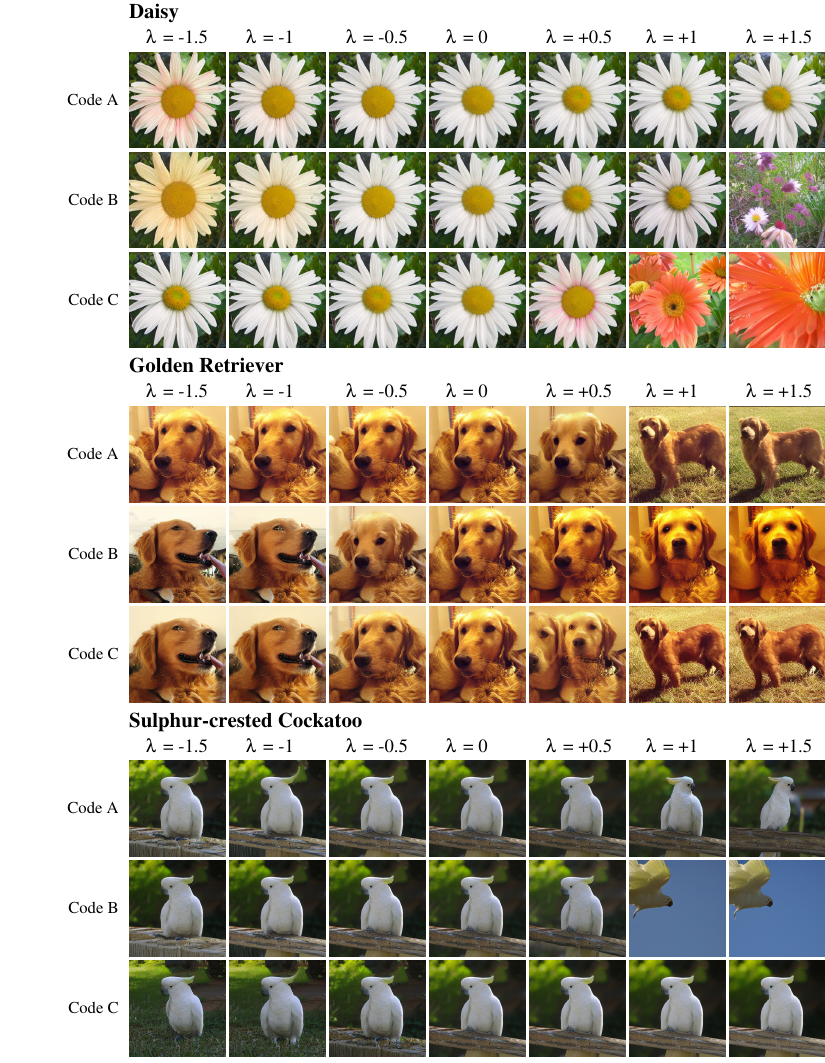}
  \caption{\textbf{Positive and negative auxiliary-conditioning scales on additional classes.}
  The layout and sampling settings follow Figure~\ref{fig:supp-lambda-sign-qualitative-1}.}
  \label{fig:supp-lambda-sign-qualitative-3}
\end{figure}
\clearpage

\clearpage

\section{Computational Cost}
\label{sec:supp-cost}
\subsection{Training Cost and Scaling with the Number of Candidates}
\label{sec:supp-training-cost}

We report the training budget, model-core computation, and measured
training throughput and GPU memory usage of WTA post-training for SiT-XL/2.

\subsubsection{Training budget and model-core computation}
\label{sec:supp-training-cost-accounting}

For each training example, we evaluate $K$ candidate predictions without
retaining their gradient graphs, select the lowest-loss candidate, and
recompute its prediction with gradients before backpropagation. Candidate
scoring is performed in chunks, where the chunk size specifies the number
of candidate--example pairs processed together. Smaller chunks limit
activation memory, while larger chunks can improve throughput through
more efficient batching. Chunking changes how candidate evaluations are
grouped, without changing their total number.

Let $F_0$ and $F_1$ denote the forward multiply--accumulate (MAC) counts
per example for ordinary SiT-XL/2 and its auxiliary-conditioned counterpart.
At global batch size $B$, using two FLOPs per MAC and a backward-to-forward
computation ratio $\beta$, the model-core FLOPs per update are
\begin{align}
 C_{\mathrm{FM}} &= 2B(1+\beta)F_0, &
 C_{\mathrm{ours}}(K) &= 2B(K+1+\beta)F_1, \\
 R(K) &= \frac{C_{\mathrm{ours}}(K)}{C_{\mathrm{FM}}}
       = \frac{K+1+\beta}{1+\beta}\frac{F_1}{F_0}.
 \label{eq:supp-training-cost-ratio}
\end{align}
Following the analytical convention used by DiT~\citep{peebles2023dit},
we use $\beta=2$. Counting MACs in the SiT-XL/2 implementation gives
$F_0=118.621$ GMAC and $F_1=118.848$ GMAC per example. The count includes
the auxiliary-conditioning projections and excludes VAE encoding,
elementwise operations, optimizer operations, and communication.
At $K=8$, the model-core computation per update is approximately
$3.674\times$ that of ordinary FM.

\paragraph{Ten-epoch training budget.}
\label{sec:supp-training-cost-k8}
At global batch size 256, ten epochs correspond to 50,040 optimizer
updates. The original SiT-XL/2 pretraining schedule uses seven million
updates at the same batch size~\citep{ma2024sit}. Post-training uses
$0.715\%$ of this update budget. Accounting for candidate computation gives
\begin{equation}
 \frac{C_{\mathrm{post}}}{C_{\mathrm{pre}}}
 = \frac{50{,}040}{7{,}000{,}000}\,R(8)
 \approx 0.0263.
 \label{eq:supp-training-cost-pretraining}
\end{equation}
Thus ten-epoch post-training at $K=8$ requires approximately
\textbf{2.63\% of the original pretraining model-core computation}.
\Cref{tab:supp-training-cost-budget} summarizes the training budget.

\begin{table}[htbp]
 \centering
 \caption{Training budget for SiT-XL/2. The pretraining schedule follows
 \citet{ma2024sit}. Model-core computation uses the analytical FLOP
 accounting in \cref{eq:supp-training-cost-ratio}.}
 \label{tab:supp-training-cost-budget}
 \begin{tabular}{lrr}
 \toprule
 Quantity & Original pretraining & $K=8$ post-training \\
 \midrule
 Training epochs & --- & 10 \\
 Optimizer updates & 7,000,000 & 50,040 \\
 Global batch size & 256 & 256 \\
 Example presentations & 1,792,000,000 & 12,810,240 \\
 Relative update budget & $100\%$ & $0.715\%$ \\
 Relative model-core computation & $100\%$ & $2.63\%$ \\
 \bottomrule
 \end{tabular}
\end{table}
\FloatBarrier

\subsubsection{Measured training efficiency}
\label{sec:supp-training-cost-scaling}

We measure training throughput on eight NVIDIA H100 80GB GPUs with global
batch size 256, using FP32 with TF32 enabled. For each $K$, we select the
highest-throughput configuration from the tested microbatch, accumulation,
and candidate-chunk settings, subject to a peak reserved-memory limit of
70 GiB per GPU. Ordinary FM uses the same microbatch choices. All reported
configurations use a per-GPU microbatch of 32 without gradient accumulation.
At $K=1$, we retain auxiliary conditioning and use the same candidate
evaluation and winner recomputation procedure.

We repeat each measurement twice, with 50 warmup updates followed by
three blocks of 50 timed updates per run. We report
the median throughput across the six blocks. We time complete training updates, including data loading
and transfer, online VAE encoding, forward and backward computation,
optimizer updates, and EMA. Checkpoint saving and preview generation
are excluded. Throughput is aggregated across all
eight GPUs using the slowest rank's elapsed time, and peak reserved memory
is the maximum across ranks.

\begin{table}[htbp]
 \centering
 \caption{Training throughput and peak reserved GPU memory on eight
 H100 80GB GPUs, using the highest-throughput tested configuration for
 each method. Throughput is reported across all eight GPUs, and memory
 per GPU.}
 \label{tab:supp-training-cost-profile}
 \begin{tabular}{lrrr}
 \toprule
 Method & Candidate chunk & Images/s $\uparrow$ & GiB/GPU \\
 \midrule
 Ordinary FM & --- & 834.51 & 36.55 \\
 Ours, $K=1$ & 32 & 656.86 & 39.66 \\
 Ours, $K=2$ & 64 & 581.01 & 39.67 \\
 Ours, $K=4$ & 128 & 474.36 & 39.66 \\
 Ours, $K=8$ & 256 & 343.75 & 39.66 \\
 Ours, $K=16$ & 512 & 224.12 & 40.96 \\
 Ours, $K=32$ & 512 & 129.79 & 40.93 \\
 \bottomrule
 \end{tabular}
\end{table}
\FloatBarrier

Training throughput decreases as $K$ increases, while chunked candidate
evaluation keeps peak reserved memory within approximately 41 GiB per GPU
across the reported configurations (\cref{tab:supp-training-cost-profile}).
Inference does not require evaluating all $K$ candidates. The number of
model evaluations per sampling step depends on the guidance configuration.
\FloatBarrier

\cpar{Measured runtime.} Importantly, the analytical FLOP overhead does not translate directly into wall-clock overhead. We benchmark all training methods under a common eight-H200 protocol. Our optimized implementation requires 0.499714 s/update, compared with 0.204752 s/update for the original FM implementation, corresponding to a $2.441\times$ measured runtime overhead, substantially below the $3.674\times$ analytical FLOP ratio. The resulting ten-epoch post-training cost is 55.57 H200 GPU-hours of projected active training.
\subsection{Computational Cost Comparison with Alternative Post-Training Methods}
All costs follow a common eight-H200 benchmark, with $B=55.57$ H200 GPU-hours corresponding to ten epochs of our post-training. To make cross-paper schedules comparable, we measure active model training at global batch 256 under common precision, synchronization, and timing boundaries. We exclude environment setup, queueing, data download, checkpoint serialization, and evaluation. CAFM and MixFlow use the authors' released implementations and official schedules. Their total GPU-hours below are projections from measured steady-state per-update times, not hardware costs reported by the original papers.

\begin{table}[t]
    \centering
    \begin{minipage}{0.92\linewidth}
        \centering
        \caption{\textbf{Projected official-schedule costs.} Active-training costs obtained by combining each released schedule with per-update timings from the common eight-H200 protocol.}
        \label{tab:native-training-costs}
        \vspace{3pt}
        \small
        \setlength{\tabcolsep}{8pt}
        \begin{tabular}{@{}l l r r@{}}
            \toprule
            Method & Schedule & GPU-h & / $B$ \\
            \midrule
            Ours & 10 ep. & 55.57 & 1.000 \\
            CAFM & 47,647 D + 2,353 G & 121.60 & 2.188 \\
            MixFlow & 200 ep. & 454.05 & 8.171 \\
            \bottomrule
        \end{tabular}
    \end{minipage}

    \vspace{12pt}

    \begin{minipage}{0.96\linewidth}
        \centering
        \caption{\textbf{Per-update costs and matched allocations.} Timings use the common H200 protocol. Here, $B=55.56823$ GPU-hours.}
        \label{tab:matched-training-costs}
        \vspace{3pt}
        \small
        \setlength{\tabcolsep}{7pt}
        \begin{tabular}{@{}lrrrr@{}}
            \toprule
            Method & TFLOPs & s/update & / Ours & Updates @ $B$ \\
            \midrule
            FM & 182.20 & 0.204752 & 0.410 & 122,127 \\
            Ours & 669.35 & 0.499714 & 1.000 & 50,040 \\
            MixFlow & -- & 0.204158 & 0.409 & 122,482 \\
            CAFM & -- & \shortstack[r]{D: 1.126702 \\ G: 0.440858} & \shortstack[r]{D: 2.255 \\ G: 0.882} & \shortstack[r]{D: 21,903 \\ G: 744} \\
            \bottomrule
        \end{tabular}
    \end{minipage}
\end{table}

\cpar{CAFM.} The CAFM paper reports that ImageNet SiT-XL/2 post-training requires approximately $4.8\times$ more wall-clock time per epoch than FM, attributing this overhead to the additional discriminator and its forward/backward Jacobian--vector product (JVP) computation \citep{lin2026continuous}. Because the paper does not specify hardware or a directly comparable timing boundary, we treat this number as context rather than using it to derive our H200 cost. We instead profile the released official implementation under the common protocol above. A discriminator update takes $t_D=1.126702$ seconds and a generator update takes $t_G=0.440858$ seconds. The released 50,000-step schedule uses a 10,000-step discriminator-only warm-up followed by a $16{:}1$ D/G update ratio. The released training schedule consists of 47,647 discriminator updates and 2,353 generator updates. Thus,
\[
T_{\mathrm{CAFM}}=47{,}647t_D+2{,}353t_G=54{,}721.29\text{ node-seconds},
\]
which is 15.20 hours on eight H200s, or 121.60 H200 GPU-hours. Under the same timing boundary, this is $5.345\times$ the original-FM active-training cost. Equivalently, CAFM requires $2.188B$, where $B$ denotes the cost of our ten-epoch post-training stage. At the matched budget $B$, we follow the official procedure and allocate 21,903 discriminator and 744 generator updates.
\cpar{MixFlow.} The MixFlow paper reports its main SiT-XL ImageNet result after 200 epochs of post-training \citep{li2026mixflow}, but does not report directly comparable H200 GPU-hours. We therefore combine the reported schedule with the authors' native released implementation measured under our common protocol. An update takes $t_{\mathrm{MixFlow}}=0.204158$ seconds, comparable to standard FM. At global batch 256, the 200-epoch endpoint contains $200\times5{,}004=1{,}000{,}800$ updates, giving
\[
C_{\mathrm{MixFlow}}=\frac{8\times1{,}000{,}800\times0.204158}{3600}=454.05\text{ H200 GPU-hours}=8.171B.
\]
Under the matched budget $B$, the authors' native implementation allows 122,482 updates (approximately 24.5 epochs).
\Cref{tab:native-training-costs} summarizes the released training schedules of the compared methods, including their projected total H200 GPU-hours and costs relative to our ten-epoch post-training. \Cref{tab:matched-training-costs} reports the measured per-update costs and the number of training updates each method receives under the same budget $B$.

Notably, the released CAFM and MixFlow training schedules involve larger additional computational budgets than our ten-epoch post-training stage. Thus, our reported high-coverage advantage is not attributable to allocating more post-training computation to our approach. We focus on these methods because they support post-training from a pretrained generator, allowing a controlled comparison of incremental training costs.

\clearpage

\end{document}